\documentclass[10pt,twocolumn,letterpaper]{article}

\usepackage{cvpr}
\usepackage{times}
\usepackage{epsfig}
\usepackage{graphicx}
\usepackage{amsmath}
\usepackage{amssymb}
\usepackage[dvipsnames]{xcolor}
\usepackage{enumitem}
\newcommand{\minisection}[1]{\vspace{0.04in} \noindent {\bf #1}\ \ }
\usepackage{algorithmicx,algorithm}
\usepackage[bold]{hhtensor}
\usepackage{capt-of}
\usepackage{multirow}
\usepackage{enumitem}
\usepackage[export]{adjustbox}
\usepackage{tabularx}
\usepackage{booktabs}  % 
\usepackage{tabu}

\cvprfinalcopy % *** Uncomment this line for the final submission
\newcommand*{\affmark}[1][*]{\textsuperscript{#1}}

\usepackage[pagebackref=true,breaklinks=true,letterpaper=true,colorlinks,bookmarks=false]{hyperref}

\def\cvprPaperID{6291} % *** Enter the CVPR Paper ID here

\ifcvprfinal\fi
\begin{document}

%%%%%%%%% TITLE
\title{Semi-supervised Learning for Few-shot Image-to-Image Translation}

\author{Yaxing Wang\affmark[1]\thanks{Work done as an intern at Inception Institute of Artificial Intelligence}, Salman Khan\affmark[2], Abel Gonzalez-Garcia\affmark[1],
Joost van de Weijer\affmark[1], Fahad  Shahbaz Khan\affmark[2,3]\\
\affmark[1]~Computer Vision Center, Universitat Aut\`onoma de Barcelona, Spain\\
\affmark[2]~Inception Institute of Artificial Intelligence, UAE \;  \affmark[3]~CVL, Link\"oping University, Sweden\\
{\tt\small \{yaxing,agonzalez,joost\}@cvc.uab.es, salman.khan@inceptioniai.org, fahad.khan@liu.se}
}

\maketitle

%%%%%%%%% ABSTRACT
\begin{abstract}
In the last few years, unpaired image-to-image translation has witnessed remarkable progress.
Although the latest methods are able to generate realistic images, they crucially rely on a large number of labeled images.  
Recently, some methods have tackled the challenging setting of few-shot image-to-image translation, reducing the labeled data requirements for the target domain during inference.
In this work, we go one step further and reduce the amount of required labeled data also from the source domain during training.
To do so, we propose applying semi-supervised learning via a noise-tolerant pseudo-labeling procedure.
We also apply a cycle consistency constraint to further exploit the information from unlabeled images, either from the same dataset or external.  
Additionally, we propose several structural modifications to facilitate the image translation task under these circumstances. 
Our  semi-supervised method for few-shot image translation, called \emph{SEMIT}, achieves excellent results on four different datasets using as little as 10\% of the source labels, and matches the performance of the main fully-supervised competitor using only 20\% labeled data. 
Our code and models are made public at: \url{https://github.com/yaxingwang/SEMIT}.

\end{abstract}

%%%%%%%%% BODY TEXT
\section{Introduction}

Image-to-image (I2I) translations are an integral part of many computer vision tasks. 
They include transformations between different modalities (\eg, from RGB to depth~\cite{liu2016learning}), between domains (\eg, %gray to color images~\cite{zhang2016colorful}, 
horses to zebras~\cite{zhu2017unpaired}) or editing operations (\eg, artistic style transfer~\cite{gatys2016image}). 
\begin{figure}[t]
    \centering
   \includegraphics[width=\columnwidth]{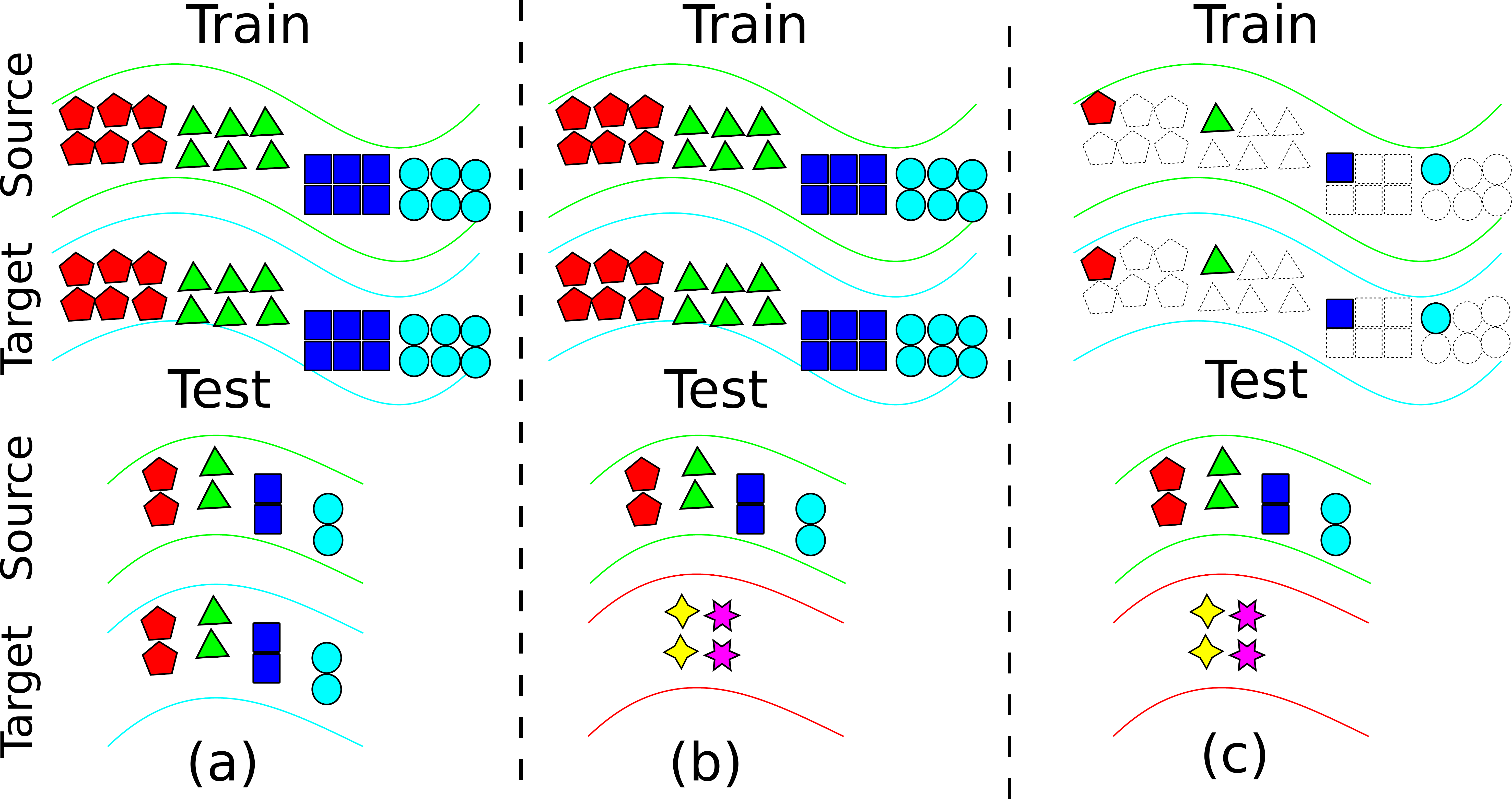}
   \caption{\small Comparison between unpaired I2I translation scenarios. Each colored symbol indicates a different image label, and dashed symbols represent unlabeled data. (a) \emph{Standard}~\cite{StarGAN2018,huang2018multimodal,zhu2017unpaired}: target classes are the same as source classes and all are seen during training. (b) \emph{Few-shot}~\cite{liu2019few}: actual target classes are different from source classes and are unseen during training. Only a few examples of the unseen target classes are available at test time. For training, source classes act temporarily as target classes.  (c) \emph{Few-shot semi-supervised} (Ours): same as few-shot, but the source domain has only a limited amount of labeled data at train time.}\vspace{-3mm}
   
   \label{fig:introduction_diagram}
\end{figure}
Benefiting from large amounts of labeled images, I2I translation has obtained great improvements on both paired~\cite{Cho_2019_CVPR,gonzalez2018image,isola2016image,wang2018mix,zhu2017toward} and unpaired image translation~\cite{Amodio_2019_CVPR,chen2019homomorphic,kim2017learning,Wu_2019_CVPR,yi2017dualgan,zhu2017unpaired}. 
Recent research trends address relevant limitations of earlier approaches, namely diversity and scalability.
Current methods~\cite{alharbi2019latent,huang2018multimodal,Lee2018drit} improve over the single-sample limitation of deterministic models by generating diverse translations given an input image. 
The scalability problem has also been successfully alleviated~\cite{StarGAN2018,perarnau2016invertible,romero2018smit,wang2019sdit}, enabling translations across several domains using a single model. 
Nonetheless, these approaches still suffer from two issues. 
First, the target domain is required to contain the same categories or attributes as the source domain at test time, therefore failing to scale to unseen categories (see Fig.~\ref{fig:introduction_diagram}(a)). 
Second, they highly rely upon having access to vast quantities of labeled data (Fig.~\ref{fig:introduction_diagram}(a, b)) at train time.
Such labels provide useful information during the training process and play a key role in some settings (\eg\ scalable I2I translation). 

Recently, several works have studied I2I translation given a few images of the \emph{target class} (as in Fig.~\ref{fig:introduction_diagram}(b)). 
Benaim and Wolf~\cite{benaim2018one} approach one-shot I2I translation by first training a variational autoencoder for the seen domain and then adapting those layers related to the unseen domain.
ZstGAN~\cite{lin2019zstgan} introduces zero-shot I2I translation, employing the annotated attributes of unseen categories instead of the labeled images.
FUNIT~\cite{liu2019few} proposes few-shot I2I translation in a multi-class setting. 
These models, however, need to be trained using large amounts of hand-annotated ground-truth labels for images of the source domain (Fig.~\ref{fig:introduction_diagram} (b)). 
Labeling large-scale datasets is costly and time consuming, making  those methods less applicable in practice. 
In this paper, we overcome this limitation and explore a novel setting, introduced in Fig.~\ref{fig:introduction_diagram}(c). 
Our focus is few-shot I2I translation %for the relevant setting 
in which only limited labeled data is available \emph{from the source classes during training}.

We propose using semi-supervised learning to reduce the requirement of labeled source images and effectively use unlabeled data. 
More concretely, we assign pseudo-labels to the unlabeled images based on an initial small set of labeled images. 
These pseudo-labels provide soft supervision to train an image translation model from source images to unseen target domains. 
Since this mechanism can potentially introduce noisy labels, 
we employ a pseudo-labeling technique that is highly robust to noisy labels. 
In order to further leverage the unlabeled images from the dataset (or even external images), we use a cycle consistency constraint~\cite{zhu2017unpaired}.
Such a cycle constraint has generally been used to guarantee the content preservation in unpaired I2I translation~\cite{kim2017learning,yi2017dualgan,zhu2017unpaired,liu2019few}, but we propose here also using it to exploit the information contained in unlabeled images. 

Additionally, we introduce further structural constraints to facilitate the I2I translation task under this challenging setting. 
First, we consider the recent Octave Convolution (OctConv) operation~\cite{chen2019drop}, which disentangles the latent representations into high and low frequency components and has achieved outstanding results for some discriminative tasks~\cite{chen2019drop}.
Since I2I translation mainly focuses on altering high-frequency information, such a disentanglement could help focalize the learning process.  
For this reason, we propose a novel application of OctConv for %eased 
I2I translation, making us the first to use it for a generative task. 
Second, we apply an effective entropy regulation procedure to make the latent representation even more domain-invariant than in previous approaches~\cite{huang2018multimodal,Lee2018drit,liu2019few}.
This leads to better generalization to target data. 
Notably, these techniques are rather generic and can be easily incorporated in many current I2I translation methods to make the task easier when there is only limited data available.

Experiments on four datasets demonstrate that the proposed method, named \emph{SEMIT}, consistently improves the performance of I2I translation using only 10\% to 20\% of the labels in the data.
Our main contributions are: 
\setlist{nolistsep}
\begin{itemize}[noitemsep]
      \item We are the first to approach few-shot I2I translation in a semi-supervised setting, reducing the amount of required labeled data for \emph{both} source and target domains.
        \item We propose several crucial modifications to facilitate this challenging setting. Our modifications can be easily adapted to other image generation architectures.
        \item We extensively study the properties of the proposed approaches on a variety of I2I translation tasks and achieve significant performance improvements.  
\end{itemize}

\section{Related work}

\begin{figure*}
    \centering
   \includegraphics[width=0.9\textwidth]{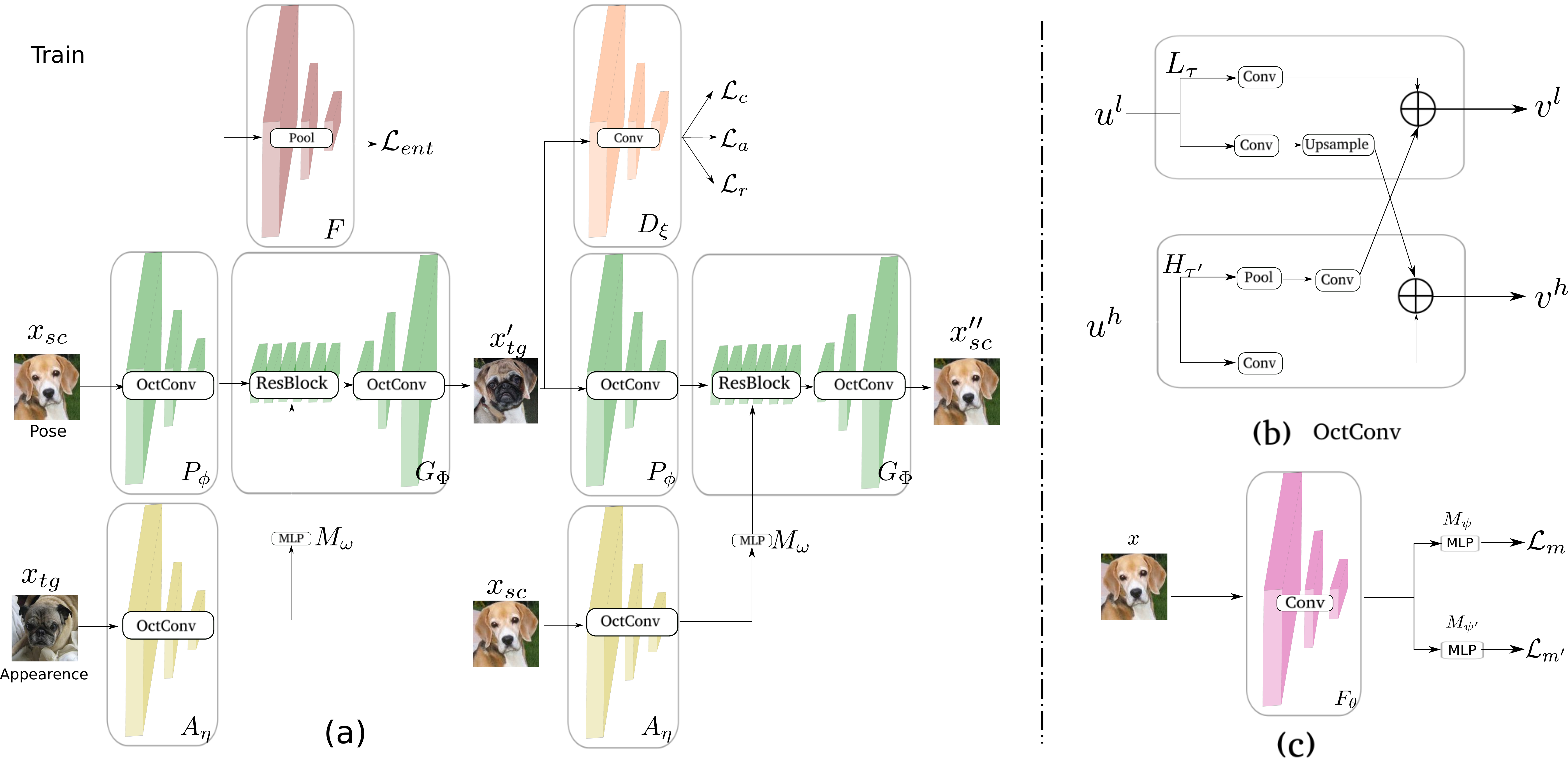}\vspace{-1mm}
   \caption{\small Model architecture for training. (a) The proposed approach is composed of two main parts:  Discriminator $D_{\xi}$ and the set of Pose encoder $P_{\phi}$,  Appearance encoder $A_{\eta}$, Generator $G_{\Phi}$, Multilayer perceptron $M_{\omega}$ and feature regulator $F$. (b) The OctConv operation contains high-frequency block ($H_{\tau'}$) and low-frequency block ($L_{\tau}$). (c) Noise-tolerant Pseudo-labeling architecture.}
   \label{fig:i2i_framework}\vspace{-3mm}
\end{figure*}
\minisection{Semi-supervised learning.} The methods in this category employ a small set of labeled images and a large set of unlabeled data to learn a general data representation.
Several works have explored applying semi-supervised learning to Generative Adversarial Networks (GANs). 
For example,~\cite{odena2016semi,salimans2016improved} merge the discriminator and classifier into a single network. 
The generated samples are used as unlabeled samples to train the ladder network~\cite{odena2016semi}.
Springenberg~\cite{springenberg2015unsupervised} explored training a classifier in a semi-supervised, adversarial manner.
Similarly, Li~\etal~\cite{chongxuan2017triple} proposed Triple-GAN that plays minimax game with a generator, a discriminator and a classifier. %three networks: the generator, the discriminator, and the classifier. 
Other works~\cite{deng2017structured,gan2017triangle} either learn two-way conditional distributions of both the labels and the images, or add a new network to predict missing labels. 
Recently, Lucic \etal~\cite{lucic2019high} proposed bottom-up and top-down methods to generate high resolution images with fewer labels. 
To the best of our knowledge, no previous work addresses I2I translation to generate highly realistic images in a semi-supervised manner. 

\minisection{Zero/few-shot I2I translation.}
Several recent works used GANs for I2I translation with few test samples. 
Lin \etal proposed zero-shot I2I translation, ZstGAN~\cite{lin2019zstgan}. 
They trained a model that separately learns domain-specific and domain-invariant features using pairs of images and captions. Benaim and Wolf~\cite{benaim2018one} instead considered one image of the target domain as an exemplar to guide image translation. % from source to target domain.  
Recently, FUNIT~\cite{liu2019few} learned a model that performs I2I translation between %multiple 
seen classes during training and scales to unseen classes during inference. %with unseen classes. 
These methods, however, rely on vast quantity of labeled source domain images  for training. 
In this work, we match their performance using only a small subset of the source domain labels. %in the source images.

\section{Proposed Approach: SEMIT}
%\YX{Keeping $\vec{x}$ and $x$ same form}
\textbf{Problem setting.} %\YX{Can we short the scetion? which is too long}
Our goal is to design an unpaired I2I translation  model that can be trained with minimal supervision (Fig.~\ref{fig:introduction_diagram} (c)). 
Importantly, in the few-shot setting the target classes are unseen during training and their few examples are made available only during the inference stage. 
In contrast to previous state-of-the-art~\cite{liu2019few}, which trains on a large number of labeled samples of the source domain (some of which act as `target' during training), we assume only limited labeled examples of the source classes are available for training.
The remaining images of the source classes are available as unlabeled examples.
Suppose we have a training set $\mathcal{D}$ with $N$  samples. One portion of the dataset is labeled, $\mathcal{D}_l = \{(\vec{x}_i,\vec{y}_i)\}_{i=1}^{N_l}$, where $\vec{x}_i \in \mathbb{R}^{D}$ denotes an image, $\vec{y}_i \in \{0,1\}^{C} : \vec{1}^\top\vec{y}_i=1$ denotes a one-hot encoded label and $C$ is the total number of classes. We consider a relatively larger unlabeled set, $\mathcal{D}_u = \{\vec{x}_i \}_{i=1}^{N_u}$, that is available for semi-supervised learning. Overall, the total number of images are $N=N_u+N_l$. %\AG{should we explicitly say that $N=N_u+N_l$?}

We initially conduct semi-supervised learning, where we learn a classifier to assign pseudo-labels $\tilde{\vec{y}}$ to the unlabeled data, generating a set $\tilde{\mathcal{D}} = \{(\vec{x}_i, \tilde{\vec{y}}_i) \}_{i=1}^{N}$, where $\tilde{\vec{y}}_i = \vec{y}_i$ for  $\vec{x}_i \in \mathcal{D}_l$ i.e., for a sample whose ground-truth label is available. 
The pseudo-labels predicted by the model form a soft label-space, i.e., $\tilde{\vec{y}}_i \in [0,1]^C : \vec{1}^\top\tilde{\vec{y}}_i =1 $. Then, our method performs unsupervised multi-domain I2I translation on the set $\tilde{\mathcal{D}}$ with few labeled images and a large unlabeled set. The dual-mode training procedure is explained below. 

\subsection{Noise-tolerant Pseudo-labeling} 
%\SK here
The assigned pseudo-labels are used to train the I2I translator network in the next stage. 
Therefore, the labeling approach must avoid generating false predictions while being able to tolerate noise in the label space. 
To achieve these requisites, we develop a Noise-tolerant Pseudo-Labeling (NTPL) approach that is trained progressively with a soft-labeling scheme to avoid the noise accumulation problem.

As illustrated in Fig.~\ref{fig:i2i_framework} (c), our pseudo-labeling scheme consists of a feature extractor $F_{\theta}$ and a couple of classification heads, $M_{\psi}$ and $M'_{\psi'}$. 
The semi-supervised labeling model is designed to suffice the following principles, (a) decision consolidation and (b) high-confidence sampling for a noise-tolerant pseudo-labeling. Firstly, the two classification heads are used to assess the uncertainty for a given unlabeled sample, i.e., a pseudo-label is considered valid only if both the classifier outputs agree with each other. Secondly, we add the pseudo-labels to the training set only if both classifier confidences are above a set threshold. Each classification head is trained using a loss $\mathcal{L}_c$ that is based on the probabilistic end-to-end noise correction framework  of~\cite{PENCIL_CVPR_2019}. 
The overall classifier loss function is the sum of losses for classification heads $M_{\psi}$ and $M'_{\psi'}$,
\vspace{-1mm}
\begin{equation}
\vspace{-2mm}
\mathcal{L}_{c} = \mathcal{L}_{m} + \mathcal{L}_{m'}.\label{eq:Lc}
\end{equation}
For both classification heads $M_{\psi}$ and $M'_{\psi'}$, %
the loss function consists of three components: (i) \textit{Compatibility loss}, which tries to match the label distribution with the pseudo-label; (ii) \textit{Classification loss}, which corrects the noise in labels; and (iii) \textit{Entropy regulation loss}, which forces the network to peak at one category rather than being flat (i.e., confusing many classes). 
Below, we explain the loss components for $\mathcal{L}_{m}$ and the formulation for loss $\mathcal{L}_{m'}$ is analogous. %

\minisection{\emph{Compatibility loss.}} The compatibility loss encourages the model to make predictions that are consistent with the ground-truth or pseudo-labels. Since in many cases, the current estimates of labels are correct, this loss function avoids estimated labels far away from the assigned labels,
\begin{align}
\mathcal{L}_{cmp} = -\frac{1}{N}\sum_{i=1}^N\sum_{j=1}^C \tilde{{y}}_{ij}\log (y^h_{ij}), 
\end{align}
where $\vec{y}^h = \text{softmax}(\vec{y}')$ is the underlying label distribution for noisy labels and $\vec{y}'$ can be updated by back-propagation during training. The tunable variable $\vec{y}'$ is initialized with $\vec{y}' = K \tilde{\vec{y}}$, where $K$ is a large scalar (1000). 

\minisection{\emph{Classification loss.}} We follow the operand-flipped KL-divergence formulation from~\cite{PENCIL_CVPR_2019}, which was shown to improve robustness against noisy labels. This loss is given by,
\begin{align}
&\mathcal{L}_{cls}  = \frac{1}{n}\sum_{i = 1}^N \text{KL}(M_{\psi}(F_{\theta}(\vec{x}_i)) ) \| \vec{y}_i^h). 
\end{align}

\minisection{\emph{Entropy regulation loss.}}  Confused models tend to output less confident predictions that are equally distributed over several object categories. The entropy regulation loss forces the estimated output distribution to be focused on one class,
\begin{eqnarray}
\label{eq:er_loss}
\resizebox{0.92\hsize}{!}{%
$\mathcal{L}_{ent} = -\frac{1}{N}\sum_{i=1}^N\sum_{j=1}^C M_{\psi}(F_{\theta}(\vec{x}_i))_j\log \big(M_{\psi}(F_{\theta}(\vec{x}_i))_j \big).$
}
\end{eqnarray}
The full loss of $M_{\psi}$  is given by, 
\begin{equation}
\label{eq:full_loss}
\mathcal{L}_{m} = \tau_{cls}\mathcal{L}_{cls} +\tau_{cmp}\mathcal{L}_{cmp} + \tau_{ent}\mathcal{L}_{ent},
\end{equation}
where $\tau _{cls}$, $\tau _{cmp}$ and $\tau_{ent}$ are the hyper-parameters. % to balance the loss contributions. 

\minisection{Training procedure.}\label{minisec: train_procedure} Our semi-supervised training procedure includes both labeled and pseudo-labeled examples. Therefore, we must select reliable pseudo-labels. 
Similar to existing work~\cite{saito2017asymmetric}, we perform the following procedure to reach this goal. Initially, we train the model (Fig.~\ref{fig:i2i_framework} (c)) with only cleanly labeled images i.e., without any pseudo-labeled images. After the sub-nets converge,  we estimate the pseudo-label for each unlabeled image $\vec{x}_i \in \mathcal{D}_u$. We define $\vec{y}^{m}_i$ and  $\vec{y}^{m'}_i$ as the predictions of $M_{\psi}$ and $M'_{\psi'}$ branches, respectively. Then, ${\ell}^{m}_i$ and ${\ell}^{m'}_i$ are the classes which have the maximum estimated probability in $\vec{y}^{m}_i$ and  $\vec{y}^{m'}_i$. We set two requirements to obtain the pseudo-label. 
First, we ensure that both the predictions agree i.e., 
${\ell}^{m}_i = {\ell}^{m'}_i$. 
At the same time, the labeling network must be highly confident about the prediction i.e., the maximum probability exceeds a threshold value (0.95). 
When both requirements are fulfilled, we assign the pseudo-label $\tilde{\vec{y}}_i$ for an unlabeled image $\vec{x}_i$. We combine both the cleanly labeled image-set and pseudo-labeled image-set to form our new training set, which is used to train the labeling network (Fig.~\ref{fig:i2i_framework} (c)). This process progressively adds reliable pseudo-labels in the training set. Besides, this cycle gradually reduces the error in the pseudo-labels for unlabeled samples. 
We repeat this process 100 times (Sec.~\ref{sec:ablation_study}).

\subsection{Unpaired Image-to-Image Translation} 
In this work, we perform unpaired I2I translation with only few labeled examples {during training}. 
Using the pseudo-labels provided by  NTPL, we now describe the actual training of the I2I translation model.

\minisection{Method overview.}
As illustrated in Fig.~\ref{fig:i2i_framework} (a), our model architecture consists of six sub-networks: Pose encoder $P_{\phi}$,  Appearance encoder $A_{\eta}$, Generator $G_{\Phi}$, Multilayer perceptron $M_{\omega}$, feature regulator $F$, and Discriminator $D_{\xi}$, where indices denote the parameters of each sub-net. 
Let $\vec{x}_{sc}  \in \mathcal{X}$ be the input source image which provides \emph{pose} information, and $\vec{x}_{tg}  \in \mathcal{X}$ the target image which contributes \emph{appearance}, with corresponding labels  $\ell_{sc}\in\left\{1,\ldots,C\right\}$ for the source and $\ell_{tg}\in\left\{1,\ldots,C\right\}$ for the target.
We use the pose extractor and the appearance extractor to encode the source %($\vec{x}_{sc}$) 
and target %($\vec{x}_{tg}$) 
images, generating $P_{\phi}(\vec{x}_{sc})$ and $A_{\eta}(\vec{x}_{tg})$, respectively. 
The appearance information $A_{\eta}(\vec{x}_{tg})$ is mapped to the input parameters of the Adaptive Instance Normalization~(AdaIN) layers~\cite{huang2018multimodal} (scale and shift) by the multilayer perceptron $M_{\omega}$. 
The generator $G_{\Phi}$ takes both the output of pose extractor $P_{\phi}(\vec{x}_{sc})$ and the AdaIN parameters output by the multilayer perceptron $M_{\omega}(A_{\eta}(\vec{x}_{tg}))$ as its input, and generates a translated output $\vec{x}'_{tg} = G_{\Phi}(P_{\phi}(\vec{x}_{sc}), M_{\omega}(A_{\eta}(\vec{x}_{tg})))$. 
We expect $G_{\Phi}$ to output a target-like image in terms of appearance, which should be classified as the corresponding label $\ell_{tg}$.   

Additionally, we generate another two images, $\vec{x}'_{sc}$ and $\vec{x}''_{sc}$, that will be used in the reconstruction loss (Eq.~\eqref{eq:recon_loss}).
The former is used to enforce content preservation~\cite{liu2019few}, and we generate it by using the source image $\vec{x}_{sc}$ as input for both the pose extractor $P_{\phi}$ and the appearance extractor $A_{\eta}$, i.e.\ $\vec{x}'_{sc} = G_{\Phi}(P_{\phi}(\vec{x}_{sc}), M_{\omega}(A_{\eta}(\vec{x}_{sc})))$\footnote{Not shown in Fig.~\ref{fig:i2i_framework} for clarity.}.
On the other hand, we generate $\vec{x}''_{sc}$ by transforming the generated target image $\vec{x}'_{tg}$ back into the source domain of $\vec{x}_{sc}$.
We achieve this by considering $\vec{x}_{sc}$ as the target appearance image, that is, $\vec{x}''_{sc} = G_{\Phi}(P_{\phi}(\vec{x}'_{tg}), M_{\omega}(A_{\eta}(\vec{x}_{sc})))$. 
This is inspired by CycleGAN~\cite{zhu2017unpaired} and using it for few-shot I2I translation is a novel application. 
The forward-backward transformation allows us to take advantage of unlabeled data since cycle consistency constraints do not require label supervision.

In order to enforce the pose features to be more class-invariant, we include an \emph{entropy regulation} loss akin to Eq.~\eqref{eq:er_loss}.
More concretely, we process input pose features via feature regulator $F$, which contains a stack of average pooling layers (hence, it does not add any parameters).
The output $F(P_{\phi}(\vec{x}_{sc}))$ is then entropy-regulated via $\mathcal{L}_{ent}$, forcing the  pose features to be sparse and focused on the overall spatial layout rather than domain-specific patterns.

A key component of our generative approach is the discriminator sub-net. 
We design the discriminator to output three terms: $D_{\xi}(\vec{x})\rightarrow \left \{ D^{c}_{\xi'}(\vec{x}), D^{a}_{\xi''}(\vec{x}), F_{\Xi}(\vec{x}) \right \}$. 
Both $D^{c}_{\xi'}\left ( \vec{x} \right ) $ and $ D^{a}_{\xi''}\left ( \vec{x} \right ) $ are probability distributions.
The goal of $D^{c}_{\xi'}\left ( \vec{x} \right )$ is to classify the generated images into their correct target class and thus guide the generator to synthesize target-specific images. We use $D^{a}_{\xi''}\left ( \vec{x} \right )$ to distinguish between real and synthesized (fake) images of the target class.  
On the other hand, $F_{\Xi}\left ( \vec{x} \right ) $ is a feature map. 
Similar to previous works~\cite{chen2016infogan,huang2018multimodal,liu2019few}, $F_{\Xi}\left ( \vec{x} \right )$ aims to match the appearance of translated image $\vec{x}'_{tg}$ to the input $\vec{x}_{tg}$.

The overall loss is a multi-task objective comprising (a) \textit{adversarial loss} that optimizes the game between the generator and the discriminator, i.e.\ $\left \{ P_{\phi}, A_{\eta}, M_{\omega}, G_{\Phi} \right \}$ seek to minimize while discriminator $D^{a}_{\xi''}$ seeks to maximize it; (b) \textit{classification loss} that ensures that sub-nets $\left \{ P_{\phi}, A_{\eta}, M_{\omega}, G_{\Phi} \right \}$ map source images $\vec{x}_{sc}$ to target-like images; (c) \textit{entropy regularization loss} that enforces the pose feature to be class-invariant; and  (d) \textit{reconstruction loss} that strengthens the connection between the translated images and the target image $\vec{x}_{tg}$,  and guarantees the translated images reserve the pose of the input source image $\vec{x}_{sc}$.

\minisection{Adversarial loss.}  We require $D^{a}_{\xi''}$ to address multiple adversarial classification tasks simultaneously, as in~\cite{liu2019few}. 
Specifically, given output $D^{a}_{\xi''} \in \mathbb{R}^{C}$, we locate the $\ell_{th}$ class response, where $\ell_{i} \in \{1,\ldots C\}$ is the category of  input image to discriminator. Using the response for $\ell_{th}$ class, we compute the adversarial loss and back-propagate gradients.  For example, when updating $D_{\xi}$,   $\ell_{th} = \ell_{sc}$; when updating $\left \{  P_{\phi}, A_{\eta}, M_{\omega}, G_{\Phi} \right \}$, $\ell_{th} \in \left \{ \ell_{sc}, \ell_{tg} \right \}$. 
We employ the following adversarial objective~\cite{goodfellow2014generative},
\begin{eqnarray}\small
 \mathcal{L}_{a} &=&  \underset{{\vec{x}_{sc}\sim \mathcal{X}}}{\mathbb{E}} \left[ \log D^{a}_{\xi''}(\vec{x}_{sc})_{\ell_{sc}} \right] \\
 &+&   \underset{\vec{x}_{sc, tg}\sim \mathcal{X}}{\mathbb{E}} \big[ \log \big(1 
 {-} D^{a}_{\xi''}(\vec{x}'_{tg})_{\ell_{tg}}\big) \big].\notag
\end{eqnarray}

\minisection{Classification loss.}
Inspired by \cite{odena2017conditional}, we use an auxiliary classifier in our GAN model to generate target-specific images. 
However, in our case the labels may be noisy for the pseudo-labeled images. 
For this reason, we employ here the noise-tolerant approach introduced in Sec.~\ref{minisec: train_procedure}
and use the single-head loss (Eq.~\eqref{eq:full_loss}) as loss function $\mathcal{L}_{c}$.
\begin{table}[t]
\setlength{\tabcolsep}{1.5pt}
\centering
\resizebox{1\columnwidth}{!}{
\begin{tabular}{|c|c|c|c|c|}
\hline
Datasets & Animals~\cite{liu2019few} & Birds~\cite{van2015building} & Flowers~\cite{nilsback2008automated} &Foods~\cite{kawano2014automatic}\\
\hline
\#classes train & 119 & 444 & 85 & 224  \\
\#classes test  & 30 & 111& 17  & 32 \\
\hline
\#images & 117,574 & 48,527& 8.189  & 31,395    \\
\hline
\end{tabular}
}
 \caption{\small Datasets used in the experiments} \vspace{-3mm}
  \label{table:datasets}%
\end{table}

\minisection{Reconstruction loss.} 
For successful I2I translation, we would like that the translated images keep the pose of the source image $\vec{x}_{sc}$ while applying the appearance of the target image $\vec{x}_{tg}$. 
We use the generated images $\vec{x}'_{sc}$ and $\vec{x}''_{sc}$ and the features $F_{\Xi}(\vec{x})$ output by the discriminator to achieve these goals via the following reconstruction loss, 
\vspace{-3pt}
\begin{equation}
\label{eq:recon_loss}
\begin{aligned}
\mathcal{L}_{r} & = \mathbb{E}_{\vec{x}_{sc}\sim \mathcal{X}, \vec{x}'_{sc} \sim \mathcal{X^{'}}}\left [\left \| \vec{x}_{sc} - \vec{x}'_{sc} \right \|_{1} \right ]\\
&+ \mathbb{E}_{\vec{x}_{sc} \sim \mathcal{X},\vec{x}''_{sc} \sim \mathcal{X}^{''}}\left [\left \| \vec{x}_{sc} - \vec{x}''_{sc} \right \|_{1} \right ]\\
&+ \mathbb{E}_{\vec{x}_{sc} \sim \mathcal{X},{\vec{x}'_{sc}} \sim \mathcal{X}^{'}} \left [\left \| F_{\Xi}(\vec{x}_{sc}) - F_{\Xi}({\vec{x}'_{sc}}) \right \|_{1} \right ]\\
&+ \mathbb{E}_{\vec{x}_{tg} \sim \mathcal{X},{\vec{x}'_{tg}} \sim \mathcal{X}^{'}} \left [\left \| F_{\Xi}(\vec{x}_{tg}) - F_{\Xi}({\vec{x}'_{tg}}) \right \|_{1} \right ].
\end{aligned}
\end{equation}

\minisection{Full Objective.}
The final loss function of our model is:
\vspace{-2pt}
\begin{align}
\label{eq:loss}
& \underset{P_{\phi},A_{\eta},M_{\omega},G_{\Phi}}{\min} \ \underset{D_{\xi}}{\max} \  \lambda_{a}\mathcal{L}_{a} +  \lambda_{c}\mathcal{L}_{c} + \lambda_{r}\mathcal{L}_{r} + \lambda_{e}\mathcal{L}_{ent},
\end{align}
where $\lambda_{a}$,  $\lambda_{c}$,   $\lambda_{r}$  and $\lambda_{e}$ are re-weighting hyper-parameters.

\begin{figure*}
    \centering
   \includegraphics[width=0.90\textwidth]{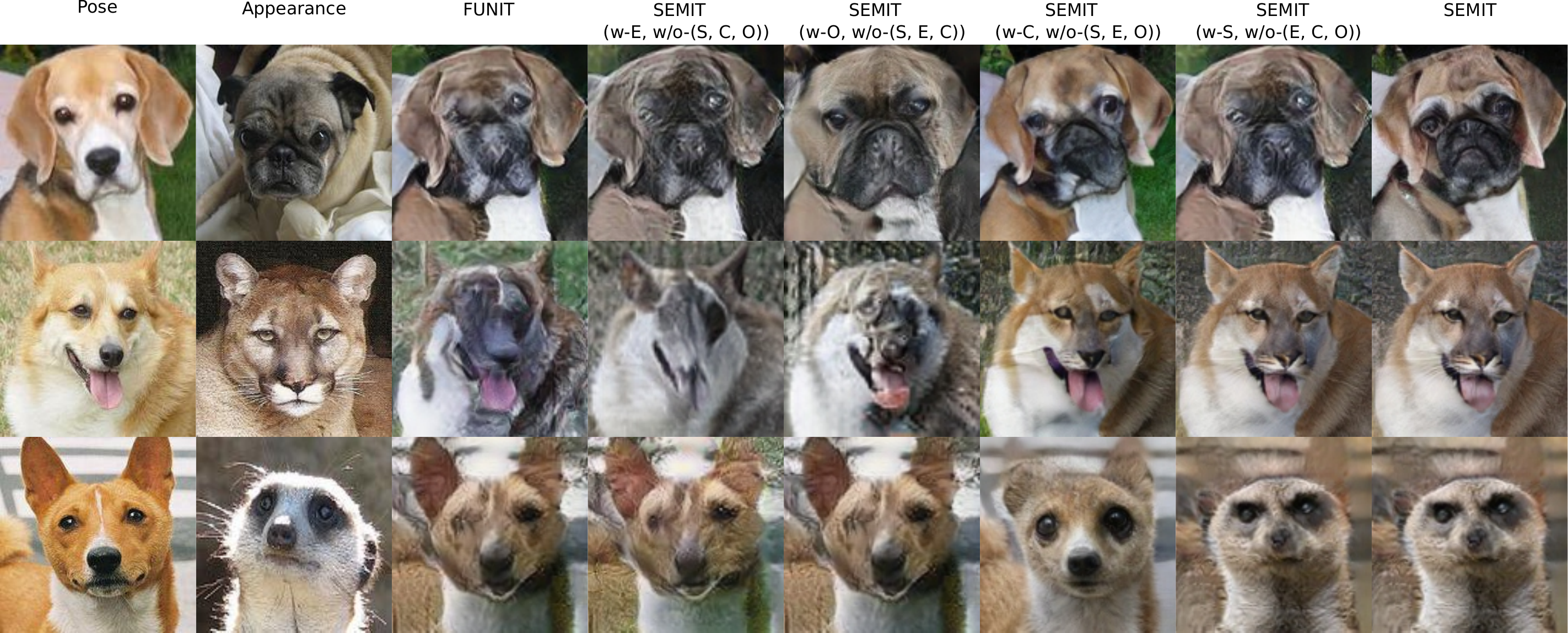}\vspace{-1mm}
   \caption{\small  Comparison between FUNIT~\cite{liu2019few} and variants of our proposed method. For example, \emph{SEMIT (w-E, w/o-(S, C, O))} indicates the model trained with only entropy regulation. More examples are in Suppl. Mat. (Sec.~1).}
   \label{fig:introduction}\vspace{-4mm}
\end{figure*}
\begin{figure}
\centering
\includegraphics[width=1\columnwidth]{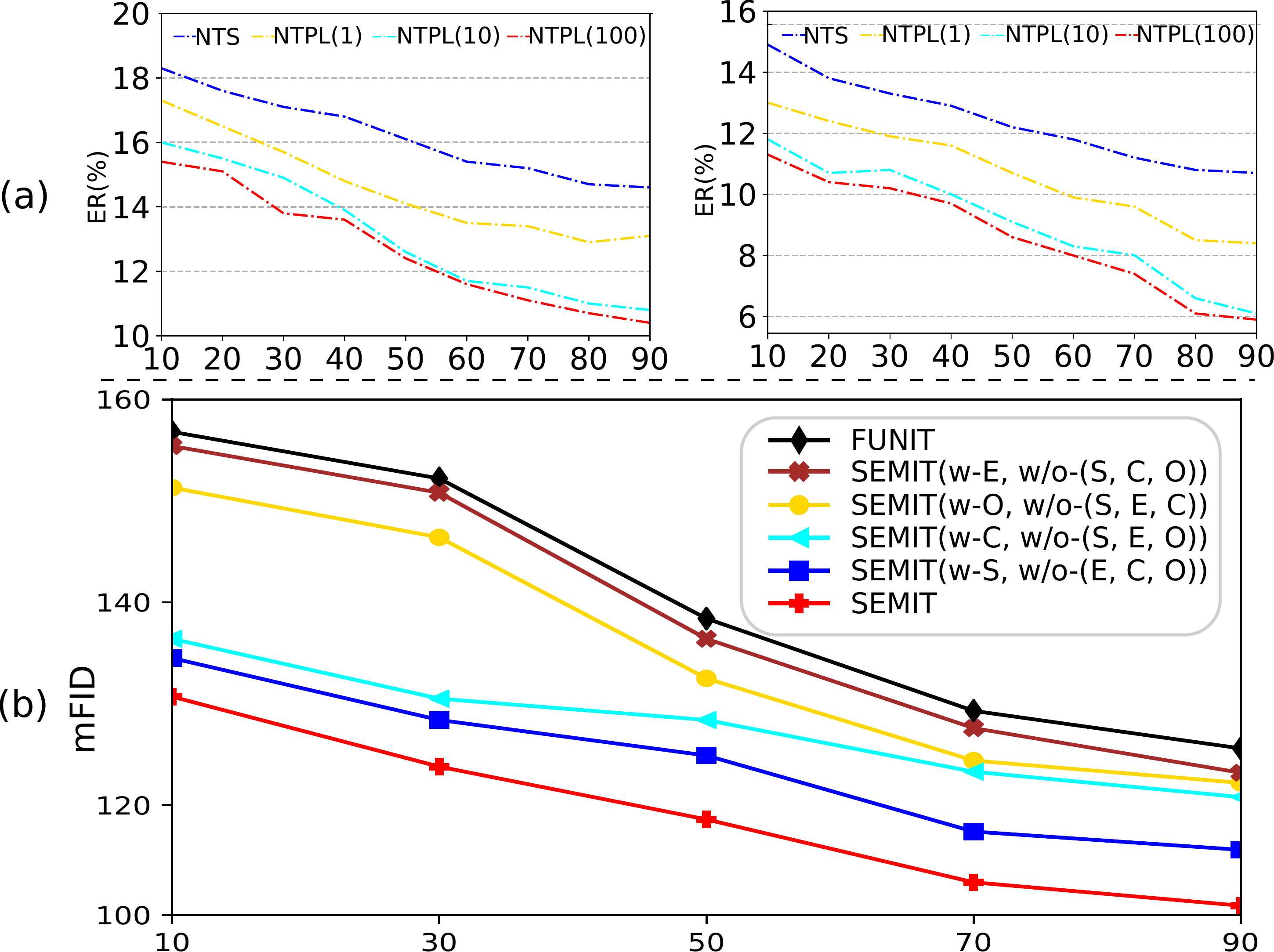}

\caption{\small (a) Ablation study on classification for (left) Animals-69 and (right) Birds, measured by Error Rate (ER). (b) Ablation study of the variants of our method for one-shot on Animals-69. The x-axis shows the percentage of the labeled data used for training.}
\label{fig:ablation_Semi-gan} 
\vspace{-3mm}
\end{figure}

\subsection{Octave network} 
\label{sec:octavenet}
An important aspect of our generator model is the Octave Convolution (OctConv) operator~\cite{chen2019drop}. 
This operator has not been studied before for generative tasks.
Specifically, OctConv aims to separate low and high-frequency feature maps. 
Since image translation mainly focuses on altering high-frequency information, such disentanglement can help with the learning. 
Furthermore, the low-frequency processing branch in OctConv layers has a wider receptive field that is useful to learn better context for the encoders.  
Let $\vec{u} = \left \{\vec{u}^h, \vec{u}^l \right \}$ and $\vec{v} = \left \{\vec{v}^h, \vec{v}^l \right \}$ be the inputs and outputs of OctConv layer, respectively. As illustrated in Fig.~\ref{fig:i2i_framework} (b), the forward pass is defined as,
\vspace{-1mm}
\begin{align}
\vspace{-1mm}
\vec{v}^l  =  L_{\tau}(\vec{u}^h, \vec{u}^l), \quad 
\vec{v}^h  =  H_{\tau'}(\vec{u}^h, \vec{u}^l), 
\label{eq:OctConv}\vspace{-3em}
\end{align}
where, $L_{\tau}$ and $H_{\tau'}$ are the high and low-frequency processing blocks with parameters $\tau$ and $\tau'$, respectively. The complete architecture of the OctConv layer  used in our work is shown in Fig.~\ref{fig:i2i_framework} (b).  
We explore suitable proportions of low-frequency and high-frequency channels for networks $ P_{\phi}$, $A_{\eta}$, and $G_{\Phi}$ in Sec.~\ref{sec:ablation_study}.
For the discriminator $D_\xi$, we empirically found the OctConv does not improve performance.

\section{Experimental setup}
\vspace{-2mm}
\minisection{Datasets.} 
We consider four datasets for evaluation, namely \emph{Animals}~\cite{liu2019few}, \emph{Birds}~\cite{van2015building}, \emph{Flowers}~\cite{nilsback2008automated}, and \emph{Foods}~\cite{kawano2014automatic} (see Table~\ref{table:datasets} for details). 
We follow FUNIT's inference procedure~\cite{liu2019few} and randomly sample 25,000 source images from the training set and translate them to each target domain (not seen during training).
We consider the 1, 5, and 20-shot settings for the target set. 
For efficiency reasons, in the ablation study we use the same smaller subset of 69 Animals categories used in~\cite{liu2019few}, which we refer to as \textit{Animals-69}.

\minisection{Evaluation metrics.}  We consider the  following three metrics. 
Among them, two are commonly used \emph{Inception Score (IS)}~\cite{salimans2016improved} and  \textit{Fr\'echet Inception Distance (FID)}~\cite{heusel2017gans}.
Moreover, we use \emph{Translation Accuracy}~\cite{liu2019few} to evaluate whether a model is able to generate images of the target class. 
Intuitively, we measure translation accuracy by the Top1 and Top5 accuracies of two classifiers: \textit{all} and \textit{test}. 
The former is trained on both source and target classes, while the latter is trained using only target classes. 

\minisection{Baselines.}
We compare against the following baselines (see Suppl. Mat. (Sec.~3) for training details).
\emph{CycleGAN}~\cite{zhu2017unpaired} uses two pairs of domain-specific encoders and decoders, trained to optimize both an adversarial loss and the cycle consistency.
\emph{StarGAN}~\cite{StarGAN2018} performs scalable image translation for all classes by inputting the label to the generator. 
\emph{MUNIT}~\cite{huang2018multimodal} disentangles the latent representation into the content space shared between two classes, and the class-specific style space.
\emph{FUNIT}~\cite{liu2019few} is the first few-shot I2I translation method.

\minisection{Variants.}
We explore a wide variety of configurations for our approach, including: semi-supervised learning (S), OctConv (O), entropy regulation (E), and cycle consistency (C). 
We denote them by \emph{SEMIT} followed by the present (w) and absent (w/o) components, \eg\ \emph{SEMIT}(w-O, w/o-(S, E, C)) refers to model with OctConv and without semi-supervised learning, entropy regulation or cycle consistency. 

\section{Experiments}

\begin{figure}[t]
\setlength{\tabcolsep}{1pt}
\centering
	\def\arraystretch{0.5}
    \begin{adjustbox}{max width=0.95\textwidth}

    \begin{tabular}{m{0.5\textwidth}m{0.5\textwidth}}
    \includegraphics[width=0.5\textwidth]{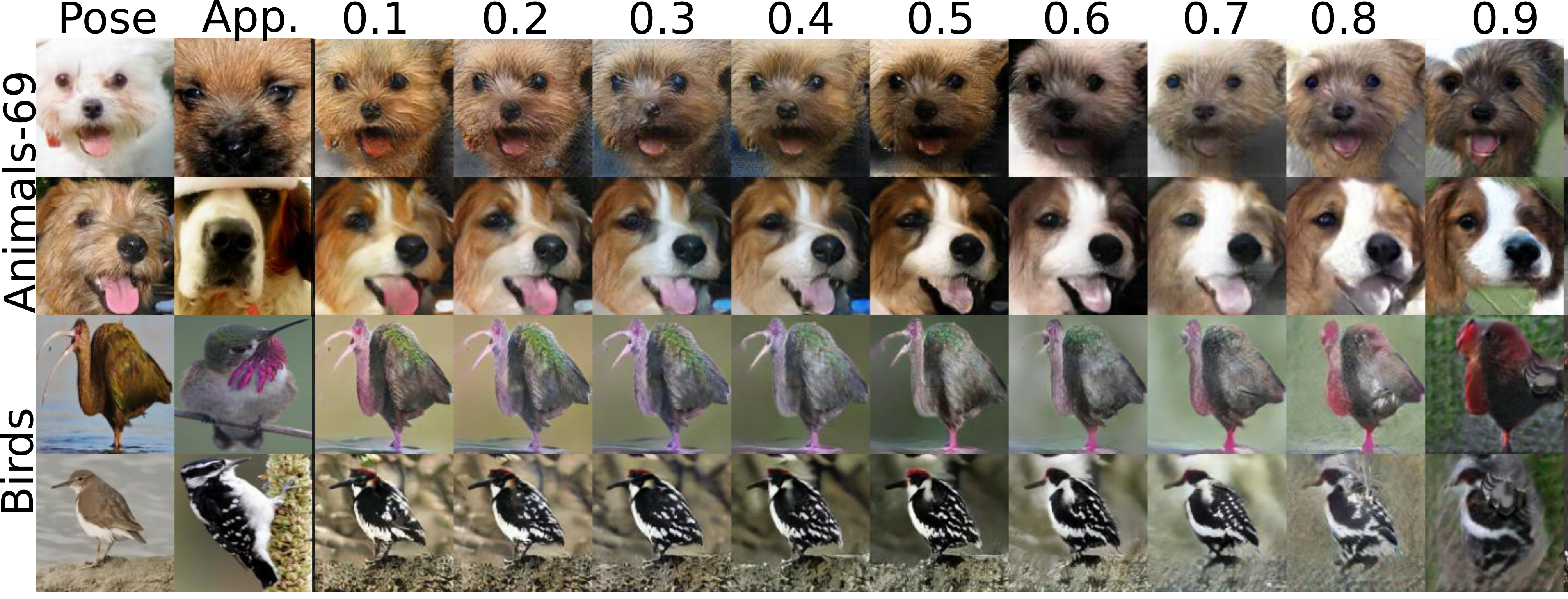}& \\  
    {\setlength{\tabcolsep}{6pt}\renewcommand{\arraystretch}{1.2}
    \resizebox{1\linewidth}{!}{
   \begin{tabular}{|c|c|c|c|c|c|c|c|c|c|}
\hline
Dataset & 0.1 & 0.2 & 0.3 & 0.4 & 0.5 & 0.6 & 0.7 & 0.8 & 0.9\\
\hline
Animals & 130.3 & 129.8 & 128.4 & 128.6 & \textbf{127.1} & 128.5 & 128.6 & 129.4 & 130.9 \\
%\hline
Birds  & 118.9 & 116.4 & 113.4 & 113.6 & \textbf{112.7} & 114.6 & 115.2 & 119.7 & 135.4   \\
\hline
\end{tabular}
            }}
    \end{tabular} 
    \end{adjustbox}
    \vspace{0mm}
    \caption{\small Qualitative (top) and quantitative
    (bottom) results for several ratios of high/low frequency channels in OctConv. Results correspond to one-shot I2I translation on Animals-69 and Birds with \textit{90\%} labeled data. More examples in Suppl. Mat. (Sec.~2).}
	\label{fig:ablation_octconv}\vspace{-3mm}
\end{figure}

\vspace{-1mm}
\subsection{Ablation study}\label{sec:ablation_study}
\vspace{-2mm}
Here, we evaluate the effect of each independent contribution to SEMIT and their combinations. Full experimental configurations are in Suppl. Mat. (Sec.~4).

\minisection{Noise-tolerant Pseudo-labeling.} 
As an alternative to our NTPL, we consider the state-of-the-art approach for fine-grained recognition NTS~\cite{yang2018learning}, as it outperforms other fine-grained methods~\cite{Lam_2017_CVPR,chen2019destruction,ge2019weakly} on our datasets.
We adopt NTS's configuration for Animals-69 and Birds and divide the datasets into train set (90\%) and test set (10\%). 
In order to study the effect of NTPL for limited labeled data, we randomly divide the train set into labeled data and unlabeled data, for which we ignore the available labels. 
All models are evaluated on the test set. 
To confirm that the iterative process in Sec.~\ref{minisec: train_procedure} leads to better performance, we consider three NTPL variants depending on the number of times that we repeat this process. 
NTPL (100) uses the standard 100 iterations to progressively add unlabeled data into the train set, whereas NTPL (10) uses 10 and NTPL (1) uses a single iteration.
We report results in terms of the Error Rate (ER) in Fig.~\ref{fig:ablation_Semi-gan} (a). %
We can see how for both NTPL and NTS, the performance is significantly lower for regimes with less labeled data. 
With 10\% of labeled data, NTS obtains a higher error than NTPL (100), \eg for Animals-69: 18.3\% vs. 15.2\%.
The training times for each variant are as follows NTS: 28.2min,  NTPL (1): 36.7min, NTPL (10): 91.2min, NTPL (100): 436min. Note that each model NTPL ($k$) is initialized with the previous model, NTPL ($k-1$).
For any given percentage of labeled data, our NTPL-based training clearly obtains superior performance, confirming that NTPL contributes to predicting better labels for unlabeled data and improves the robustness against noisy labels.

\begin{table}[t]%bth!
    \centering
    \resizebox{0.95\linewidth}{!}{\mbox{%
	{\tabcolsep=1pt\def\arraystretch{1}
		\begin{tabular}{|c|c|cccc|cc|c|}
		\hline
    % 		\hline
			& Setting & Top1-all  &  Top5-all  & Top1-test &  Top5-test &  IS-all  &  IS-test &  mFID \\
			\hline
			\parbox[t]{2mm}{\multirow{8}{*}{\rotatebox[origin=c]{90}{\small{100\%}}}}
			& \texttt{\small{CycleGAN-20}}&  28.97&  47.88&  38.32&  71.82&  10.48&  7.43&  197.13\\
			& \texttt{MUNIT-20}&  38.61&  62.94&  53.90&  84.00&    10.20&  7.59&  158.93\\
% 			\cline{2-10}
			& \texttt{StarGAN-20}&  24.71&  48.92&  35.23&  73.75&   8.57&  6.21&  198.07 \\
% 			\cline{2-10}
			& \texttt{FUNIT-1}&  17.07&  54.11&  46.72&  82.36&    22.18&  10.04&  93.03 \\
			& \texttt{FUNIT-5}&  33.29&  78.19&  68.68&  96.05&    22.56&  13.33&  70.24 \\
			& \texttt{FUNIT-20}&  { 39.10}&  { 84.39}&  { 73.69}&  { 97.96}&    { 22.54}&  {14.82}&  { 66.14}\\
			& \texttt{SEMIT-1}&  { 29.42}&  { 65.51}&  { 62.47}&  { 90.29}&  { 24.48}&  { 13.87}&  { 75.87}\\
			& \texttt{SEMIT-5}&  { 35.48}&  { 78.96}&  { 71.23}&  { 94.86}&    { 25.63}&  { 15.68}&  { 68.32}\\
			& \texttt{SEMIT-20}&  {\bf 45.70}&  {\bf 88.5}&  {\bf 74.86}&  {\bf 99.51}&    {\bf 26.23}&  {\bf 16.31}&  {\bf 49.84}\\
			\hline
			\parbox[t]{2mm}{\multirow{6}{*}{\rotatebox[origin=c]{90}{\small{20\%}}}}
			& \texttt{FUNIT-1}&  { 12.01}&  { 30.59}&  { 29.86}&  { 55.44}&   { 19.23}&  {4.59}&  { 139.7}\\
			& \texttt{FUNIT-5}&  { 15.25}&  { 36.48}&  { 36.47}&  { 66.58}&   { 21.12}&  {6.16}&  { 128.3}\\
			& \texttt{FUNIT-20}&  { 16.95}&  { 41.43}&  { 42.61}&  { 68.92}&  { 21.48}&  {6.78}&  { 117.4}\\
			& \texttt{SEMIT-1}&  { 26.71}&  {  69.48}&  { 65.48}&  { 85.49}&    { 23.52}&  { 12.63}&  { 92.21}\\
			& \texttt{SEMIT-5}&  { 39.56}&  {  78.34}&  { 71.81}&  { 96.25}&   { 24.01}&  { 14.17}&  { 69.28}\\
			& \texttt{SEMIT-20}&  {\bf 44.25}&  { \bf 85.60}&  {\bf 73.80}&  {\bf 98.62}&   {\bf 24.67}&  {\bf 15.04}&  {\bf 65.21}\\
			\hline
			\parbox[t]{2mm}{\multirow{6}{*}{\rotatebox[origin=c]{90}{\small{10\%}}}}
			& \texttt{FUNIT-1}&  { 10.21}&  { 28.41}&  { 27.42}&  { 49.54}&    { 17.24}&  {4.05}&  { 156.8}\\
			& \texttt{FUNIT-5}&  { 13.04}&  { 35.62}&  { 31.21}&  { 61.70}&   { 19.12}&  {4.87}&  { 138.8}\\
			& \texttt{FUNIT-20}&  { 14.84}&  { 39.64}&  { 37.52}&  { 65.84}&   { 19.64}&  {5.53}&  { 127.8}\\
			& \texttt{SEMIT-1}&  {16.25}&  {51.55}&  {39.71}&  {81.47}&    {22.58}& { 8.61}&  {99.42}\\
			& \texttt{SEMIT-5}&  {29.40}&  {76.14}&  {62.72}&  {92.13}&    {22.98}& { 13.24}&  {78.46}\\
			& \texttt{SEMIT-20}&  \textbf{39.02}&  \textbf{82.90}&  \textbf{69.70}&  \textbf{95.40}&   \textbf{23.43}& \textbf{ 14.07}&  \textbf{69.40}\\
			\hline
		\end{tabular}}%
		}}\vspace{1mm}
    	\caption{Performance comparison with baselines on Animals~\cite{liu2019few}.}\label{tab: results_animals}\vspace{-4mm}
	\end{table}

\minisection{OctConv layer.}
Fig.~\ref{fig:ablation_octconv} (top) presents qualitative results on the Animals-69 and Birds datasets (one-shot, 90\% labeled data) for varying proportions of channels devoted to high or low frequencies (Sec.~\ref{sec:octavenet}).
Changing this value has a clear effect on how our method generates images. 
As reported in Fig.~\ref{fig:ablation_octconv} (bottom), we find using OctConv with half the channels for each frequency (0.5) obtains the best performance. For the rest of the paper, we set this value to 0.5. 
We conclude that OctConv facilitates the I2I translation task by disentangling the feature space into frequencies.

\begin{figure*}
    \centering
    \includegraphics[width=0.68\textwidth,angle=90]{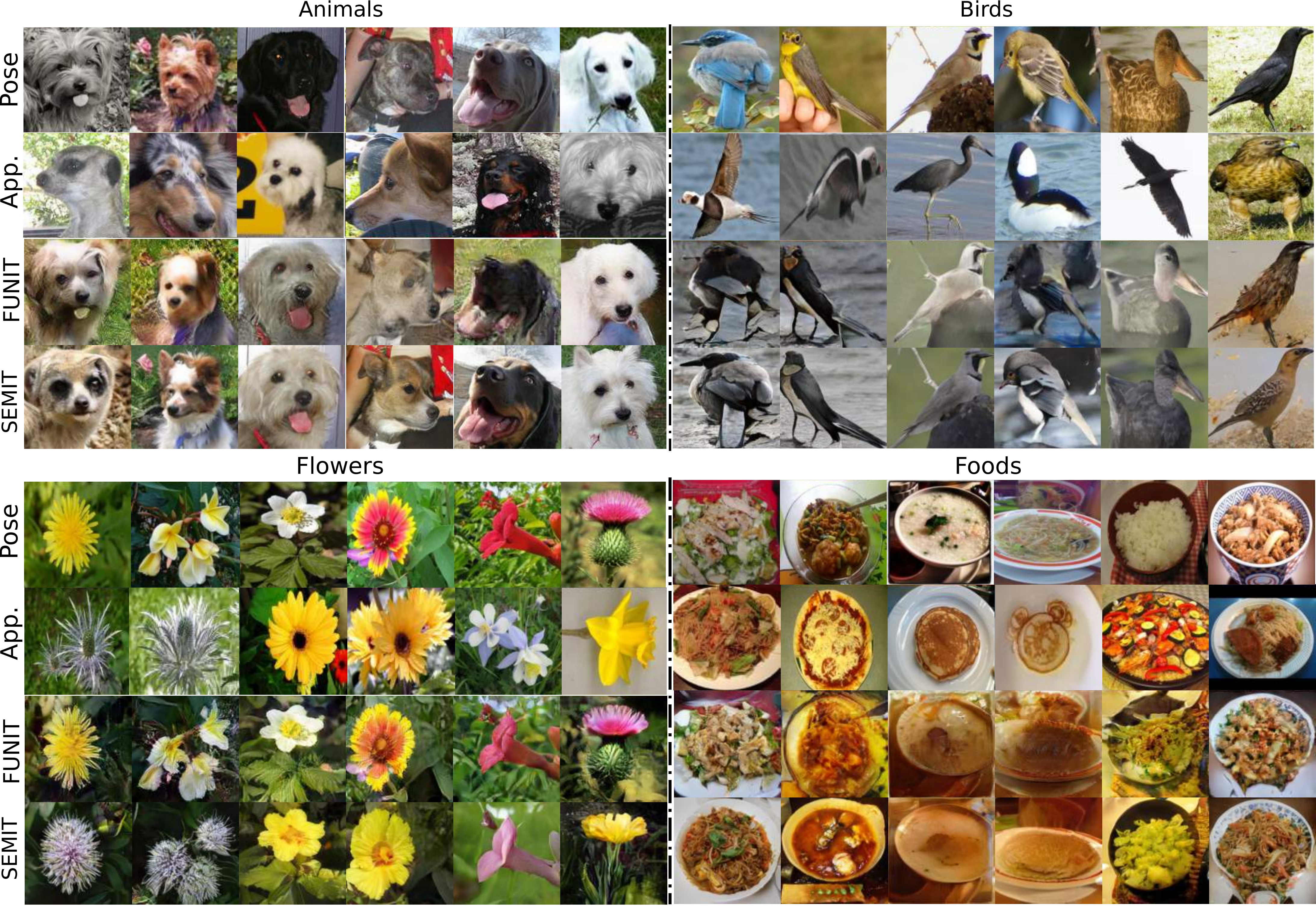}\vspace{-1mm}
   \caption{\small Qualitative comparison between our method and FUNIT~\cite{liu2019few} on the four datasets. More examples are in Suppl. Mat. (Sec.~6).}\vspace{-3mm}
%      Best viewed electronically with zoom.
   \label{fig:i2i_comparasion}
\end{figure*}

\minisection{Other SEMIT variants.} 
Fig.~\ref{fig:ablation_Semi-gan} (b) presents a comparison between several variants of SEMIT and FUNIT~\cite{liu2019few} in terms of mean FID (mFID) for various percentages of labeled training data. 
Adding either Entropy regulation (SEMIT (w-E, w/o-(S, C, O))
or OctConv layers (SEMIT (w-O, w/o-(S, E, C)) improves the performance of I2I translation compared to FUNIT~\cite{liu2019few} at all levels of labeled data.
We attribute this to the architectural advantage and enhanced optimization granted by our contributions to the I2I translation task in general.   
%which benefits from the alleviating of I2I translation task. 
Next, adding either cycle consistency or semi-supervised learning achieves a further boost in performance.
The improvement is remarkably substantial for
%the regimes with
low percentages of labeled data (10\%-30\%), which is our main focus. This shows how such techniques, especially semi-supervised learning, can truly exploit the information  in unlabeled data and thus relax the labeled data requirements.
Finally, the complete SEMIT obtains the best mFID score, indicating that our method successfully performs I2I translation even with much fewer labeled images.
Similar conclusions can be drawn from the qualitative examples in Fig.~\ref{fig:introduction}, where SEMIT successfully transfers the appearance of the given target to the input pose image.

\begin{table}[t]%bth!
    \centering
    \resizebox{0.95\linewidth}{!}{\mbox{%
	{\tabcolsep=1pt\def\arraystretch{1}
		\begin{tabular}{|c|c|cccc|cc|c|}
			\hline
			& Setting & Top1-all  &  Top5-all  & Top1-test &  Top5-test &   IS-all  &  IS-test &  mFID \\
			\hline
			\parbox[t]{2mm}{\multirow{8}{*}{\rotatebox[origin=c]{90}{\small{100\%}}}}
			& \texttt{CycleGAN-20}&  9.24&  22.37&  19.46&  42.56&   25.28&  7.11&  215.30\\
			& \texttt{MUNIT-20}&  23.12&  41.41&  38.76&  62.71&    24.76&  9.66&  198.55\\
% 			\cline{2-10}
			& \texttt{StarGAN-20}&  5.38&  16.02&  13.95&  33.96&  18.94&  5.24&  260.04\\
% 			\cline{2-10}
			& \texttt{FUNIT-1}&  11.17&  34.38&  30.86&  60.19&   67.17&  17.16&  113.53\\
			& \texttt{FUNIT-5}&  20.24&  51.61&  45.40&  75.75&  74.81&  22.37&  99.72\\
			& \texttt{FUNIT-20}&  {23.50}&  {56.37}&  {49.81}&    {1.286}&  76.42& { 24.00}&  {97.94}\\
			& \texttt{SEMIT-1}&  { 15.64}&  {  42.85}&  { 43.7.62}&  { 72.41}&    { 69.63}&  { 20.12}&  { 105.82}\\
			& \texttt{SEMIT-5}&  {23.57}&  {  55.96}&  { 49.42}&  {80.41}&   {78.42}&  { 24.98}&  { 90.48}\\
			& \texttt{SEMIT-20}&  {\bf 28.15}&  { \bf 62.41}&  {\bf 54.62}&  {\bf 83.32}&  {\bf 82.64}&  {\bf 27.51}&  {\bf 83.56}\\
			\hline
			\parbox[t]{2mm}{\multirow{6}{*}{\rotatebox[origin=c]{90}{\small{20\%}}}}
			& \texttt{FUNIT-1}&  { 6.21}&  { 20.31}&  { 15.34}&  { 28.45}&   { 29.23}&  {8.23}&  { 184.4}\\
			& \texttt{FUNIT-5}&  { 10.25}&  { 22.34}&  { 22.75}&  { 43.24}&   { 43.62}&  {12.53}&  { 168.6}\\
			& \texttt{FUNIT-20}&  { 11.76}&  { 28.51}&  { 26.47}&  { 46.38}&   { 58.40}&  {15.75}&  { 145.1}\\
			& \texttt{SEMIT-1}&  { 13.58}&  {  48.16}&  { 43.97}&  {64.27}&   {59.29}&  { 16.48}&  { 109.84}\\
			& \texttt{SEMIT-5}&  {19.23}&  {  53.25}&  {50.34}&  { 73.16}&   { 67.84}&  { 22.27}&  { 98.38}\\
			& \texttt{SEMIT-20}&  {\bf 21.49}&  { \bf 57.55}&  {\bf 52.34}&  {\bf 76.41}&   {\bf 72.31}&  {\bf 23.44}&  {\bf 95.41}\\
			\hline
			\parbox[t]{2mm}{\multirow{6}{*}{\rotatebox[origin=c]{90}{\small{10\%}}}}
			& \texttt{FUNIT-1}&  { 6.04}&  { 19.34}&  { 12.51}&  { 38.84}&  { 32.62}&  {7.47}&  { 203.3}\\
			& \texttt{FUNIT-5}&  { 8.82}&  { 22.52}&  { 19.85}&  { 42.53}&   { 38.59}&  {9.53}&  { 175.7}\\
			& \texttt{FUNIT-20}&  { 10.98}&  { 26.41}&  { 22.48}&  { 48.36}&   { 41.37}&  {13.85}&  { 154.9}\\
			& \texttt{SEMIT-1}&  {11.21}& {37.14}&  {35.14}&  {59.41}& {48.48}& { 12.57}&  {128.4}\\
			& \texttt{SEMIT-5}&  {13.54}&  {43.63}&  {40.24}&  {68.75}&  {59.84}&{ 17.58}&  {119.4}\\
			& \texttt{SEMIT-20}&  \textbf{15.41}&  \textbf{48.36}&  \textbf{42.51}&  \textbf{71.49}&    \textbf{65.42}& \textbf{ 19.87}&  \textbf{109.8}\\
			\hline
		\end{tabular}}%
		}}\vspace{1mm}
    	\caption{Performance comparison with  baselines on Birds~\cite{van2015building}.}\label{tab: results_birds}\vspace{-3mm}
	\end{table}

\subsection{Results for models trained on a single dataset}\label{sec:results_single dataset}
	
Tables~\ref{tab: results_animals} and~\ref{tab: results_birds} report results for all baselines and our method on Animals~\cite{liu2019few} and Birds~\cite{van2015building}, under three percentages of labeled source images: 10\%, 20\%, and 100\%.
We use the 20-shot setting as default for all baselines but also explore 1-shot and 5-shot settings for FUNIT~\cite{liu2019few} and our method. 
All the baselines that are not specialized for few-shot translation (i.e.\ CycleGAN~\cite{zhu2017unpaired}, MUNIT~\cite{zhu2017toward}, and StarGAN~\cite{StarGAN2018}) suffer a significant disadvantage in the few-shot scenario, obtaining %much worse 
inferior results even with 100\% of labeled images.
However, both FUNIT and SEMIT perform significantly better, and SEMIT achieves the %absolute 
best results for all metrics under all settings. 
Importantly, SEMIT trained with only 20\% of ground-truth labels (\eg\ mFID of 65.21 for Animals) is comparable to FUNIT with 100\% labeled data (mFID 66.14),  clearly indicating that the proposed method effectively performs I2I translation with $\times5$ less labeled data.     
Finally, our method achieves competitive performance even with only 10\% available labeled data.  We also provide many-shot case in Suppl. Mat. (Sec. ~5)

Fig.~\ref{fig:i2i_comparasion} shows example images generated by FUNIT and SEMIT using 10\% labeled data. 
On Animals, Birds, and Food, FUNIT manages to generate somewhat adequate target-specific images. 
Nonetheless, under closer inspection, the images look blurry and unrealistic, since FUNIT fails to acquire enough guidance for generation without exploiting the information present in unlabeled data. 
Besides, it completely fails to synthesize target-specific images of Flowers, possibly due to the smaller number of images per class in this dataset.  
SEMIT, however, successfully synthesizes convincing target-specific images for all datasets, including the challenging Flowers dataset. These results again support our conclusion: SEMIT effectively applies the target appearance onto the given pose image despite using much less labeled data.

\subsection{Results for models trained on multiple datasets}\label{sec:results_multiple dataset}
\begin{figure}[t]
\centering
\includegraphics[width=0.80\columnwidth]{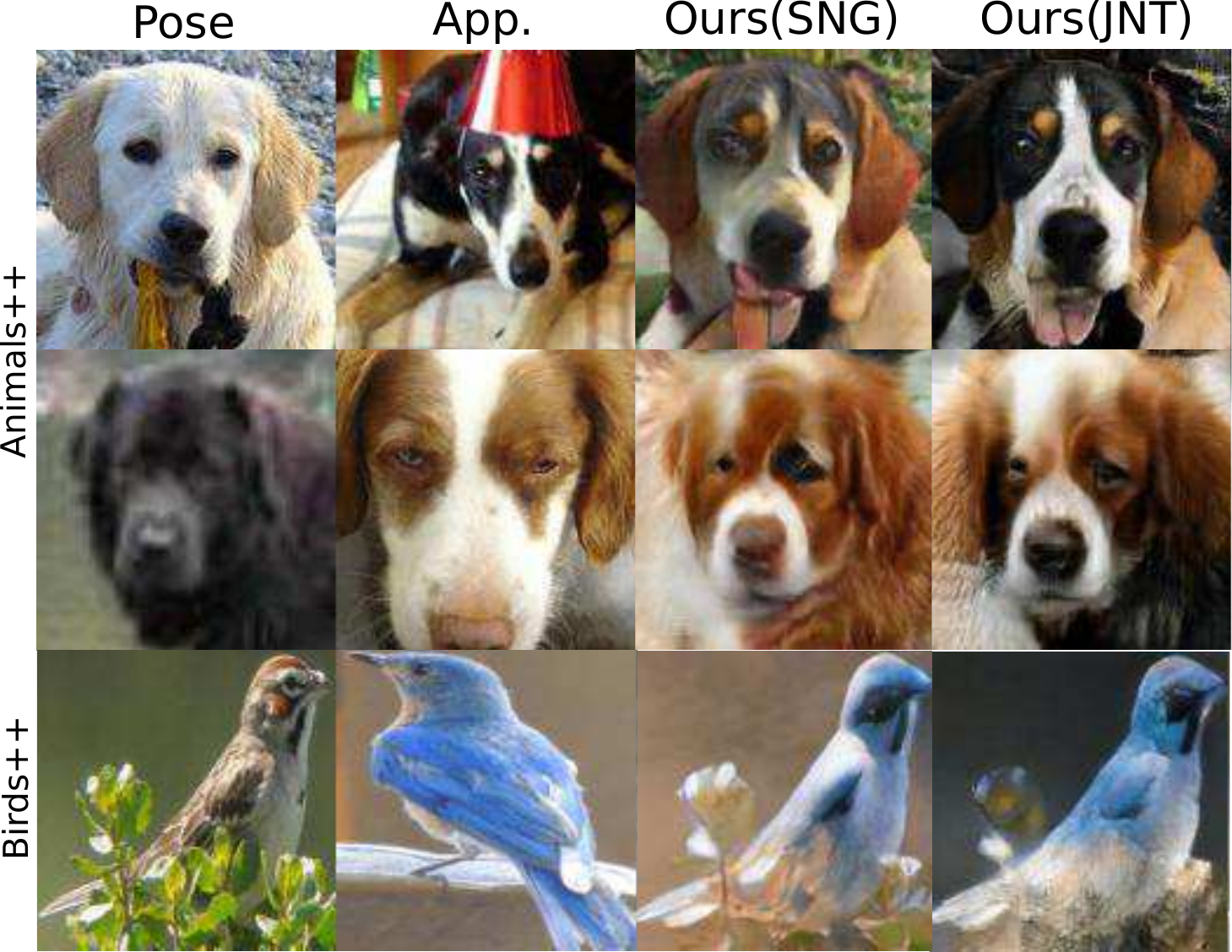}
\caption{\label{fig:results_multidatasets} \small Results of our method on a single dataset (SNG) and joint datasets (JNT). More examples are in Suppl. Mat. (Sec.~7).}
\vspace{-3mm}
\end{figure}

\vspace{-2mm}
We investigate whether SEMIT can learn from multiple datasets simultaneously.
For this, we merge an additional 20,000 unlabeled animal faces (from~\cite{Lee2018drit,zhang2008cat,KhoslaYaoJayadevaprakashFeiFei_FGVC2011} or retrieved via search engine) into the Animals dataset, which we call  \textit{Animals++}.
We also combine 6,033 unlabeled bird images from CUB-200-2011~\cite{welinder2010caltech} into Birds and name it \textit{Birds++}. 
We term our model trained on the original dataset as Ours (SNG)  and  the model trained using the expanded versions as Ours (JNT).
We experiment using 10\% labeled data from the original datasets.
Note, we do not apply the classification loss (Eq.~\ref{eq:Lc}) for the newly added images, as the external data might include classes not in the source set.
Fig.~\ref{fig:results_multidatasets} shows  results which illustrate how Ours (SNG) achieves successful target-specific I2I translation, but Ours (JNT) exhibits even higher visual quality.
% and obtains lower scores on both datasets.
This is because Ours (JNT) can leverage the additional low-level information (color, texture, etc.) provided by the additional data. We provide quantitative results in Suppl. Mat. (Sec.~8).

\section{Conclusions}
\vspace{-2mm}
We proposed semi-supervised learning to perform few-shot unpaired I2I translation with fewer image labels for the source domain. Moreover, we employ a cycle consistency constraint to exploit the information in unlabeled data, as well as several generic modifications to make the I2I translation task easier. Our method achieves excellent results on several datasets while requiring only a fraction of the labels.

\minisection{Acknowledgements.} We thank the Spanish project TIN2016-79717-R and also its CERCA Program of the Generalitat de Catalunya. 

{\small
\bibliographystyle{ieee_fullname}
%\bibliography{longstrings, egbib}
\bibliography{shortstrings, egbib}
}

\appendix
\section{SEMIT variants}
\begin{figure*}
    \centering
   \includegraphics[width=1\textwidth]{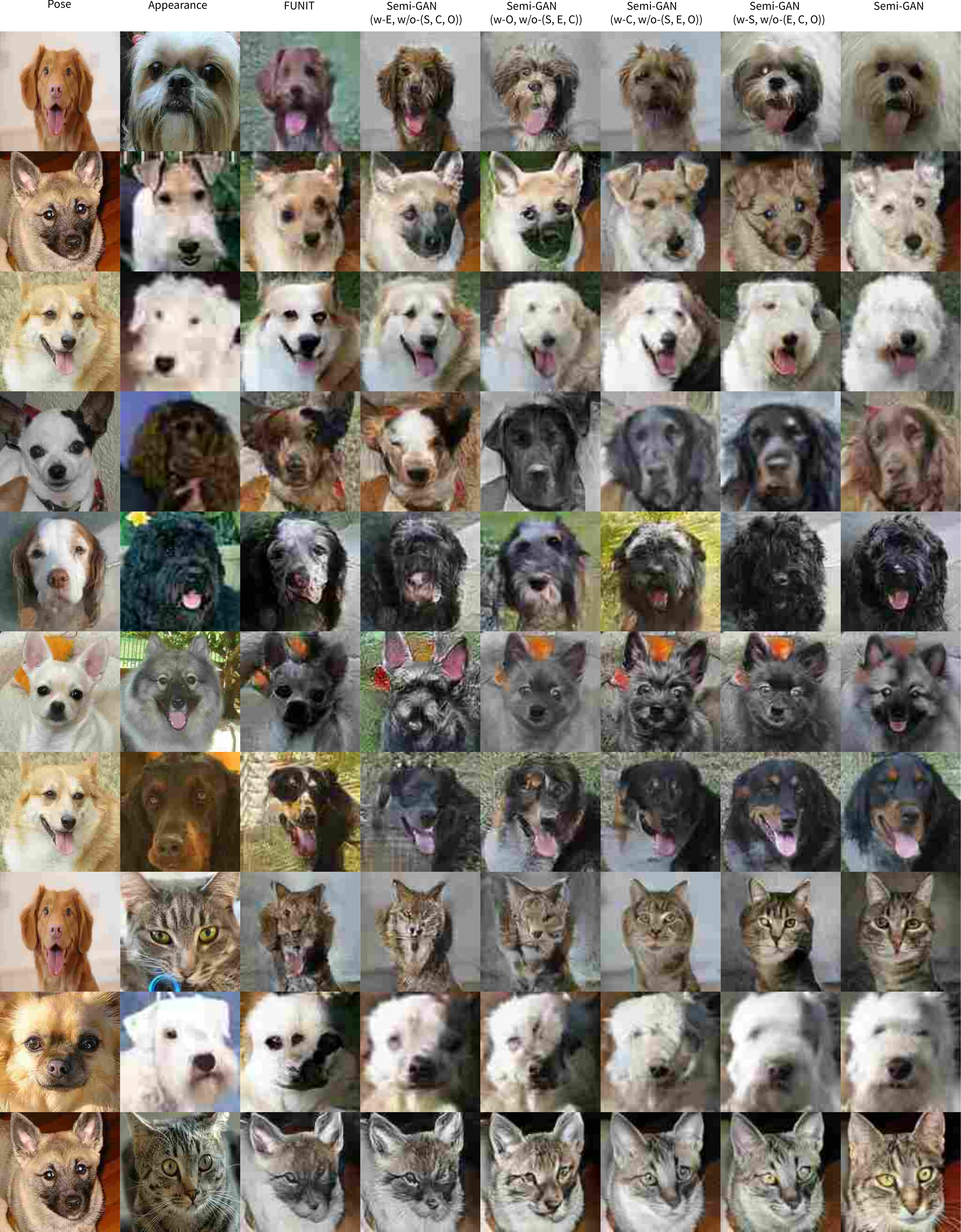}\vspace{-1mm}
   \caption{\small Comparison between FUNIT~\cite{liu2019few} and variants of our proposed method. For example, \emph{SEMIT(w-E, w/o-(S, C, O))} indicates the model trained with only entropy regulation.}
   \label{fig:Semi-GAN_variants}\vspace{-4mm}
\end{figure*}

We provide additional results for the experiments of SEMIT variants in Fig.~\ref{fig:Semi-GAN_variants}.

\section{Effect of OctConv layer}
\begin{figure*}
    \centering
   \includegraphics[width=1\textwidth]{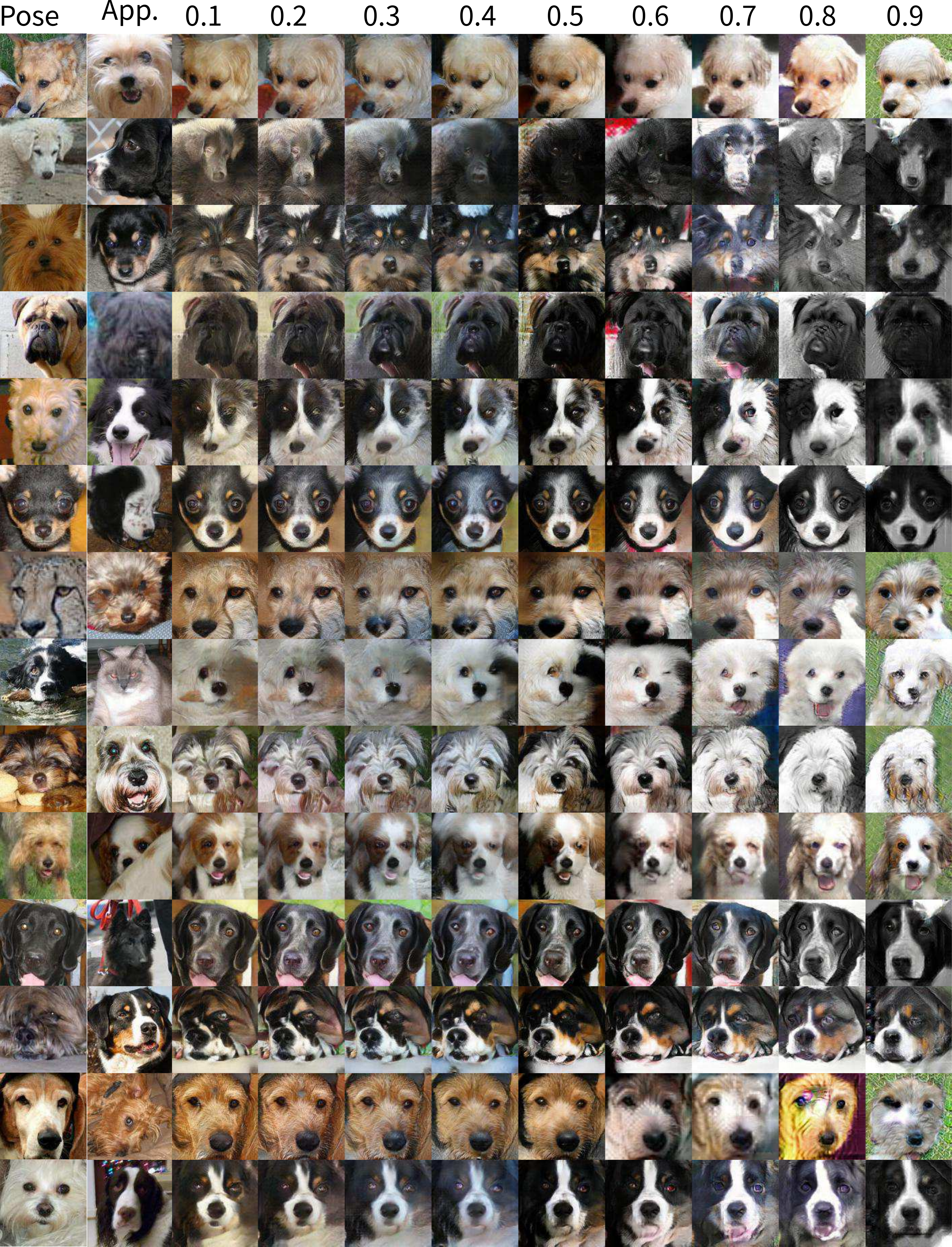}\vspace{-1mm}
   \caption{\small Qualitative results for several ratios of high/low frequency channels in OctConv. Results correspond to one-shot I2I translation on Animals-69  with 90\% labeled data.}
   \label{fig:octave_animal69}\vspace{-4mm}
\end{figure*}

\begin{figure*}
    \centering
   \includegraphics[width=0.9\textwidth]{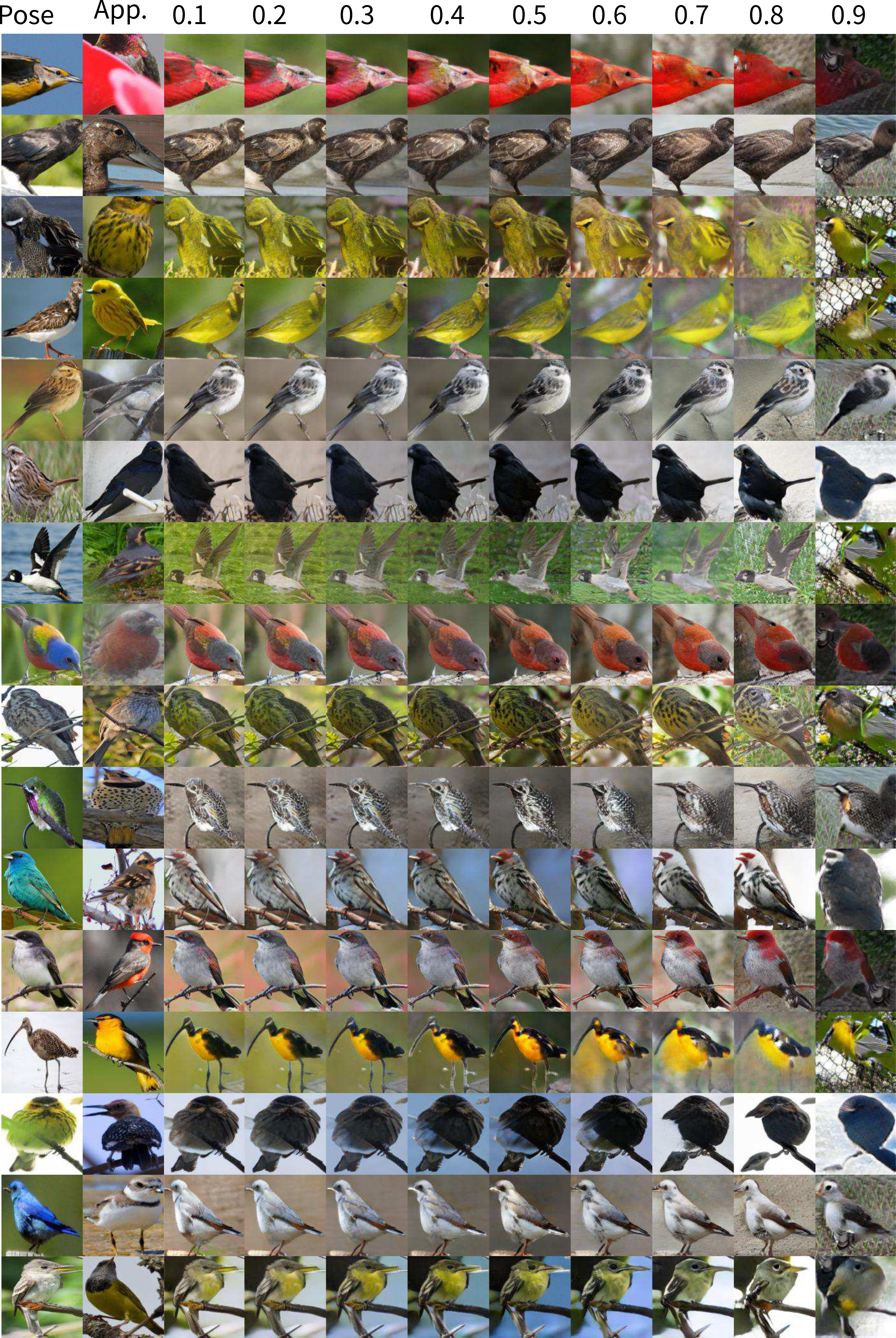}\vspace{-1mm}
   \caption{\small Qualitative results for several ratios of high/low frequency channels in OctConv. Results correspond to one-shot I2I translation on Birds with 90\% labeled data.}
   \label{fig:octave_birds}\vspace{-4mm}
\end{figure*}

We provide additional results for the experiments on studying the effect of OctConv layer in Figs.~\ref{fig:octave_animal69} and \ref{fig:octave_birds}.

\section{Baselines}

FUNIT~\cite{liu2019few} considers two training settings for the baselines, \textit{fair} and \textit{unfair}.  
Here, we only consider the unfair setting as it is more challenging. 
The unfair setting gives these baselines access (during training time) to the target images from the test data. 
In order to train CycleGAN and MUNIT, which are only able to perform I2I translation between two domains given during training, we define the source data as one domain and the target data as the other. 
Both StarGAN~\cite{StarGAN2018} and FUNIT~\cite{liu2019few} do not have this requirement as they approach multi-domain I2I translation.

\section{Training settings} 
All models are implemented in PyTorch~\cite{paszke2017automatic}. Our model architecture consists of six sub-networks: Pose encoder $P_{\phi}$,  Appearance encoder $A_{\eta}$, Generator $G_{\Phi}$, Multilayer perceptron $M_{\omega}$, Feature regulator $F$, and Discriminator $D_{\xi}$. The Pose encoder $P_{\phi}$ contains 3 OctConv layers and 6 blocks. Each OctConv layer uses $4 \times 4$ filters with stride 2, except for the first one which uses $7 \times 7$ with stride 1, and each block contains two OctConv layers with $3 \times 3$ filters and stride of 1.  The Appearance encoder $A_{\eta}$ consists of 3 OctConv layers, one average pooling layer and a $1 \times 1$ convolutional layer. $M_{\omega}$ consists of two fully connected layers with 256 and 4096 units~(8-256-4096), and takes the output of the Appearance encoder as input. The Generator $G_{\Phi}$ comprises of ResBlock layers and two fractionally strided OctConv layers. 
The ResBlock consists of 6 residual blocks, as in the Pose encoder $P_{\phi}$, but including AdaIN~\cite{huang2018multimodal} layers. 
The AdaIN layers take the output of $P_{\phi}$ and the output of  $M_{\omega}$ as input. The Feature regulator $F$, which takes the output of the Pose encoder as input, includes two average pooling layers with stride 2. For the $D_{\xi}$, we use one $7 \times 7$ convolutional layer with stride 1,  and then 4 blocks for the Feature extractor $F_{\Xi}(\vec{x})$. Each block contains two ResBlocks, one average pooling layer with stride 2, and one $1 \times 1$ convolutional layer. The Feature extractor $F_{\Xi}(\vec{x})$ is followed by two parallel sub-networks, each of them containing one convolutional layer with $3 \times 3$ filters and stride 1. The OctConv operation contains high-frequency block ($H_{\tau'}$) and low-frequency block ($L_{\tau}$). The former consists of two parallel branches: (a) the $3 \times 3$ convolutional layer with stride 1, and (b) one pooling layer and one convolutional layer which uses $3 \times 3$ with stride 1. The latter also  contains two parallel branches: (a) the $3 \times 3$ convolutional layer with stride 1, and (b) one convolutional layer which uses $3 \times 3$ with stride 1 and one upsampling layer.

We randomly initialize the weights following a Gaussian distribution, and optimize the model using RMSProp with batch size 128, with a learning rate of 0.0001. Inspired by existing methods~\cite{liu2019few,miyato2018spectral}, we employ the hinge loss variant of the GAN loss in our experiment.  For the three hyper-parameters in Eq.~5 ($\tau _{cls}$, $\tau _{cmp}$ and $\tau_{ent}$), we use the same parameter values as in~\cite{PENCIL_CVPR_2019}. %, which already conducted an ablation study to tune them for a similar problem. 
Two of the hyper-parameters in Eq.~8 ($\lambda_{a}$ and $\lambda_{r}$) are set to the values used in FUNIT~\cite{liu2019few}. This demonstrates how our method is not particularly sensitive to hyper-parameter tuning and it works with the default values.
The remaining parameters ($\lambda_{c}$ and $\lambda_{e}$) are optimized with a line search $10^p$ with $p\in\{-5,-4,...,4,5\}$. %These details will be added to the paper.  
In all experiments, we use the following hyper-parameters:  $  \lambda_{a} =1 $,   $\lambda_{c} = 0.1$,  $\lambda_{r} = 0.1$, $\lambda_{e} = 0.1$. For the experiment on noise-tolerant pseudo-labeling,  we use the following hyper-parameters: $\tau_{cls} = 1/C$, $\tau_{cmp}=0.1/C$ and $\tau_{ent}=0.8/C$, where $C$ is the number of categories.   We use 8 V100 GPUs in an NVIDIA DGX1 machine to perform all our experiments. 

\section{Quantitative results for many-shot}~\label{many-shot}
We  also provide results for \textit{many-shot} on a single dataset in  Tabs.~\ref{tab: supp_results_animals},~\ref{tab: supp_results_birds}. Given more appearance images per class, our method obtains a higher performance. But the gap between 100-shot and 20-shot is small, clearly indicating that our method already obtains good results while using only few appearance images.

\begin{table}[t]%bth!
    \centering
    \resizebox{0.95\linewidth}{!}{\mbox{%
	{\tabcolsep=1pt\def\arraystretch{1}
		\begin{tabular}{|c|c|cccc|cc|c|}
		\hline
    % 		\hline
			& Setting & Top1-all  &  Top5-all  & Top1-test &  Top5-test &  IS-all  &  IS-test &  mFID \\
			\hline
			\parbox[t]{2mm}{\multirow{3}{*}{\rotatebox[origin=c]{90}{\small{100\%}}}}
			& \texttt{SEMIT-1}&  { 29.42}&  { 65.51}&  { 62.47}&  { 90.29}&  { 24.48}&  { 13.87}&  { 75.87}\\
			& \texttt{SEMIT-5}&  { 35.48}&  { 78.96}&  { 71.23}&  { 94.86}&    { 25.63}&  { 15.68}&  { 68.32}\\
			& \texttt{SEMIT-20}&  {   45.70}&  {   88.5}&  {   74.86}&  {  \bf 99.51}&    {   26.23}&  {   16.31}&  {   49.84}\\
			& \texttt{SEMIT-100}&  {\bf 45.93}&  {\bf 89.6}&  {\bf 76.01}&  { 99.30}&    {\bf 28.54}&  {\bf 17.26}&  {\bf 47.98}\\
			\hline
			\parbox[t]{2mm}{\multirow{3}{*}{\rotatebox[origin=c]{90}{\small{20\%}}}}
			& \texttt{SEMIT-1}&  { 26.71}&  {  69.48}&  { 65.48}&  { 85.49}&    { 23.52}&  { 12.63}&  { 92.21}\\
			& \texttt{SEMIT-5}&  { 39.56}&  {  78.34}&  { 71.81}&  { 96.25}&   { 24.01}&  { 14.17}&  { 69.28}\\
			& \texttt{SEMIT-20}&  {  44.25}&  {   85.60}&  {  73.80}&  {  98.62}&   {  24.67}&  {  15.04}&  {  65.21}\\
			& \texttt{SEMIT-100}&  { \bf45.97}&  { \bf   87.58}&  {  \bf 74.59}&  { \bf  98.96}&   {\bf 26.67}&  {  \bf 16.07}&  {  \bf 64.52}\\
			\hline
			\parbox[t]{2mm}{\multirow{3}{*}{\rotatebox[origin=c]{90}{\small{10\%}}}}
			& \texttt{SEMIT-1}&  {16.25}&  {51.55}&  {39.71}&  {81.47}&    {22.58}& { 8.61}&  {99.42}\\
			& \texttt{SEMIT-5}&  {29.40}&  {76.14}&  {62.72}&  {92.13}&    {22.98}& { 13.24}&  {78.46}\\
			& \texttt{SEMIT-20}&    {39.02}&    {82.90}&    {69.70}&    {95.40}&     {23.43}&   { 14.07}&    {69.40}\\
			& \texttt{SEMIT-100}&  \textbf{39.91}&  \textbf{84.06}&  \textbf{70.58}&  \textbf{95.99}&   \textbf{24.04}& \textbf{ 14.89}&  \textbf{67.78}\\
			\hline
			\hline
		\end{tabular}}%
		}}\vspace{1mm}
    	\caption{Performance comparison with baselines on Animals~\cite{liu2019few}.}\label{tab: supp_results_animals}
	\end{table}

\begin{table}[t]%bth!
    \centering
    \resizebox{0.95\linewidth}{!}{\mbox{%
	{\tabcolsep=1pt\def\arraystretch{1}
		\begin{tabular}{|c|c|cccc|cc|c|}
			\hline
			& Setting & Top1-all  &  Top5-all  & Top1-test &  Top5-test &   IS-all  &  IS-test &  mFID \\
			\hline
			\parbox[t]{2mm}{\multirow{4}{*}{\rotatebox[origin=c]{90}{\small{100\%}}}}
			& \texttt{SEMIT-1}&  { 15.64}&  {  42.85}&  { 43.7.62}&  { 72.41}&    { 69.63}&  { 20.12}&  { 105.82}\\
			& \texttt{SEMIT-5}&  {23.57}&  {  55.96}&  { 49.42}&  {80.41}&   {78.42}&  { 24.98}&  { 90.48}\\
			& \texttt{SEMIT-20}&  {\bf 28.15}&  {  62.41}&  {54.62}&  { 83.32}&  { 82.64}&  {27.51}&  {83.56}\\
			& \texttt{SEMIT-100}&  {  27.08}&  { \bf 64.7 }&  {\bf  55.31}&  { \bf 83.80}&  { \bf 83.47}&  { \bf 27.94}&  {\bf  81.03}\\
			\hline
			\parbox[t]{2mm}{\multirow{4}{*}{\rotatebox[origin=c]{90}{\small{20\%}}}}
			& \texttt{SEMIT-1}&  { 13.58}&  {  48.16}&  { 43.97}&  {64.27}&   {59.29}&  { 16.48}&  { 109.84}\\
			& \texttt{SEMIT-5}&  {19.23}&  {  53.25}&  {50.34}&  { 73.16}&   { 67.84}&  { 22.27}&  { 98.38}\\
			& \texttt{SEMIT-20}&  {21.49}&  {  57.55}&  { 52.34}&  { 76.41}&   { 72.31}&  { \bf 23.44}&  { 95.41}\\
			& \texttt{SEMIT-100}&  {\bf 21.62}&  { \bf 57.79}&  {\bf 53.36}&  {\bf 76.97}&   {\bf 74.26}&  { 23.02}&  {\bf 93.81}\\
			\hline
			\parbox[t]{2mm}{\multirow{4}{*}{\rotatebox[origin=c]{90}{\small{10\%}}}}
			& \texttt{SEMIT-1}&  {11.21}& {37.14}&  {35.14}&  {59.41}& {48.48}& { 12.57}&  {128.4}\\
			& \texttt{SEMIT-5}&  {13.54}&  {43.63}&  {40.24}&  {68.75}&  {59.84}&{ 17.58}&  {119.4}\\
			& \texttt{SEMIT-20}&  {15.41}&   {48.36}&   {42.51}&   {71.49}&     \textbf{65.42}&  { 19.87}&   {109.8}\\
			& \texttt{SEMIT-100}&  \textbf{16.87}&  \textbf{50.68}&  \textbf{42.65}&  \textbf{73.64}&    {64.85}& \textbf{ 20.45}&  \textbf{108.1}\\
			\hline
		\end{tabular}}%
		}}\vspace{1mm}
    	\caption{Performance comparison with  baselines on Birds~\cite{van2015building}.}\label{tab: supp_results_birds}
	\end{table}	

\begin{table}[t]%bth!
    \centering
    \resizebox{0.95\linewidth}{!}{\mbox{%
	{\tabcolsep=1pt\def\arraystretch{1}
		\begin{tabular}{|c|c|cccc|cc|c|}
		\hline
    % 		\hline
			& Setting & Top1-all  &  Top5-all  & Top1-test &  Top5-test &  IS-all  &  IS-test &  mFID \\
			\hline
			\parbox[t]{2mm}{\multirow{6}{*}{\rotatebox[origin=c]{90}{\small{10\%}}}}
			& \texttt{FUNIT-1}&  { 10.21}&  { 28.41}&  { 27.42}&  { 49.54}&    { 17.24}&  {4.05}&  { 156.8}\\
			& \texttt{FUNIT-5}&  { 13.04}&  { 35.62}&  { 31.21}&  { 61.70}&   { 19.12}&  {4.87}&  { 138.8}\\
			& \texttt{FUNIT-20}&  { 14.84}&  { 39.64}&  { 37.52}&  { 65.84}&   { 19.64}&  {5.53}&  { 127.8}\\
			& \texttt{Ours(SNG)-1}&  {16.25}&  {51.55}&  {39.71}&  {81.47}&    {22.58}& { 8.61}&  {99.42}\\
			& \texttt{Ours(SNG)-5}&  {29.40}&  {76.14}&  {62.72}&  {92.13}&    {22.98}& { 13.24}&  {78.46}\\
			& \texttt{Ours(SNG)-20}&  {39.02}&  {82.90}&  {69.70}&  {95.40}&   {23.43}& { 14.07}&  {69.40}\\
			& \texttt{Ours(JNT)-1}&  {18.42}&  {53.86}&  {41.23}&  {85.59}&    {22.47}& { 10.61}&  {88.62}\\
			& \texttt{Ours(JNT)-5}&  {31.52}&  {79.25}&  {63.81}&  {94.54}&    {23.05}& { 14.83}&  {69.57}\\
			& \texttt{Ours(JNT)-20}&  \textbf{51.26}&  \textbf{83.51}&  \textbf{73.86}&  \textbf{96.74}&   \textbf{24.85}& \textbf{ 15.64}&  \textbf{62.31}\\
			\hline
		\end{tabular}}%
		}}\vspace{1mm}
    	\caption{Performance comparison with baselines on Animals++.}\label{tab: results_animals}\vspace{-3mm}
	\end{table}	

\begin{table}[t]%bth!
    \centering
    \resizebox{0.95\linewidth}{!}{\mbox{%
	{\tabcolsep=1pt\def\arraystretch{1}
		\begin{tabular}{|c|c|cccc|cc|c|}
			\hline
			& Setting & Top1-all  &  Top5-all  & Top1-test &  Top5-test &   IS-all  &  IS-test &  mFID \\
			\hline
			\parbox[t]{2mm}{\multirow{6}{*}{\rotatebox[origin=c]{90}{\small{10\%}}}}
			& \texttt{FUNIT-1}&  { 6.04}&  { 19.34}&  { 12.51}&  { 38.84}&  { 32.62}&  {7.47}&  { 203.3}\\
			& \texttt{FUNIT-5}&  { 8.82}&  { 22.52}&  { 19.85}&  { 42.53}&   { 38.59}&  {9.53}&  { 175.7}\\
			& \texttt{FUNIT-20}&  { 10.98}&  { 26.41}&  { 22.48}&  { 48.36}&   { 41.37}&  {13.85}&  { 154.9}\\
			& \texttt{Ours(SNG)-1}&  {11.21}& {37.14}&  {35.14}&  {59.41}& {48.48}& { 12.57}&  {128.40}\\
			& \texttt{Ours(SNG)-5}&  {13.54}&  {43.63}&  {40.24}&  {68.75}&  {59.84}&{ 17.58}&  {119.44}\\
			& \texttt{Ours(SNG)-20}&  {15.41}&  {48.36}&  {42.51}&  {71.49}&    {65.42}& { 19.87}&  {109.81}\\
			& \texttt{Ours(JNT)-1}&  {14.69}& {42.15}&  {36.72}&  {65.19}& {67.48}& { 18.12}&  {108.97}\\
			& \texttt{Ours(JNT)-5}&  {21.92}&  {55.86}&  {48.17}&  {76.33}&  {73.97}&{ 23.84}&  {96.23}\\
			& \texttt{Ours(JNT)-20}&  \textbf{26.23}&  \textbf{61.34}&  \textbf{52.68}&  \textbf{79.97}&    \textbf{78.31}& \textbf{ 25.49}&  \textbf{95.41}\\
			\hline
		\end{tabular}}%
		}}\vspace{1mm}
    	\caption{Performance comparison with  baselines on Birds++.}\label{tab: results_birds}\vspace{-3mm}
	\end{table}

\section{Qualitative results for models trained on a single dataset}
\begin{figure*}
    \centering
   \includegraphics[width=1\textwidth]{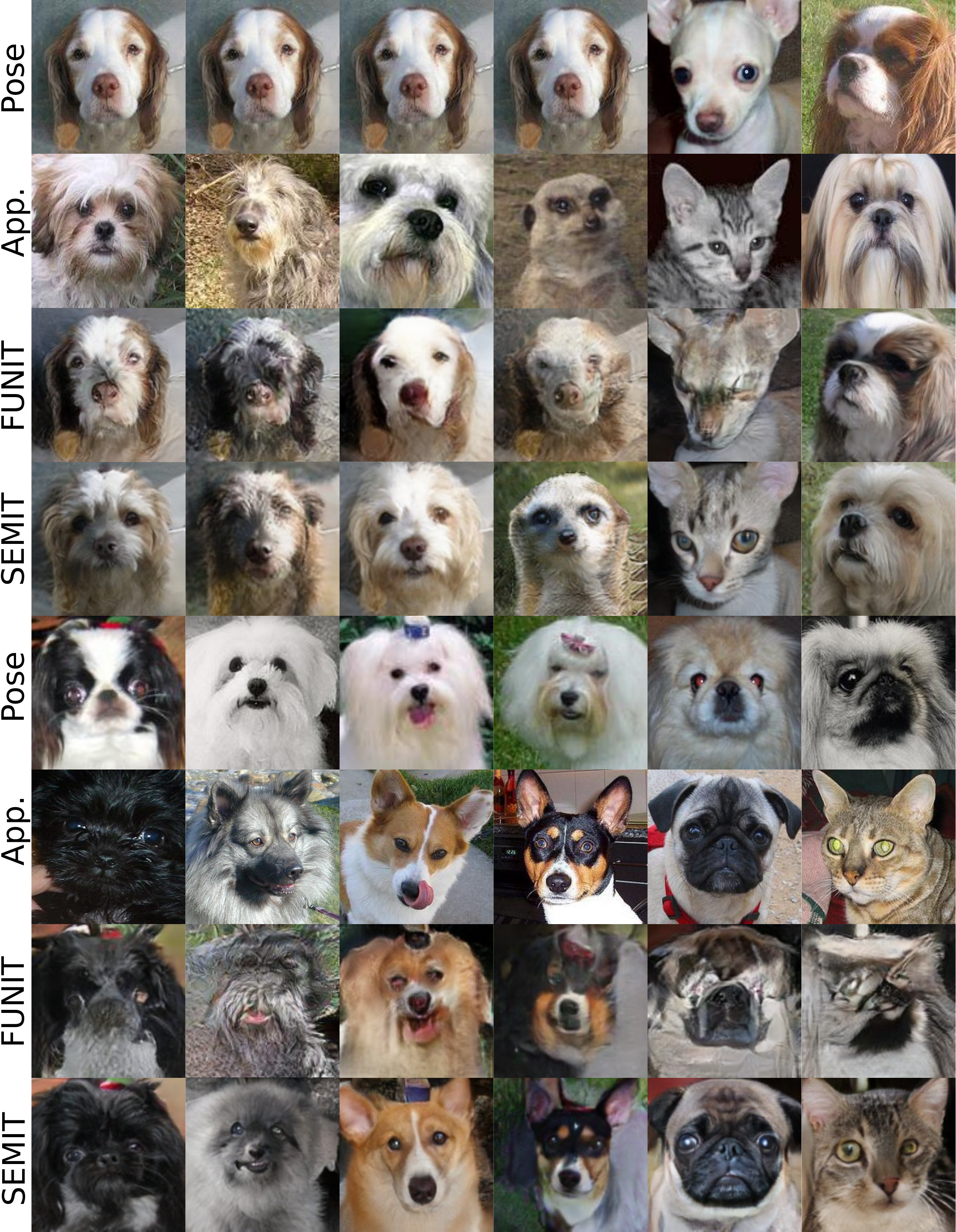}\vspace{-1mm}
   \caption{\small Qualitative comparison on the Animals datasets.}
   \label{fig:single_dataset_animals}\vspace{-4mm}
\end{figure*}

\begin{figure*}
    \centering
   \includegraphics[width=0.9\textwidth]{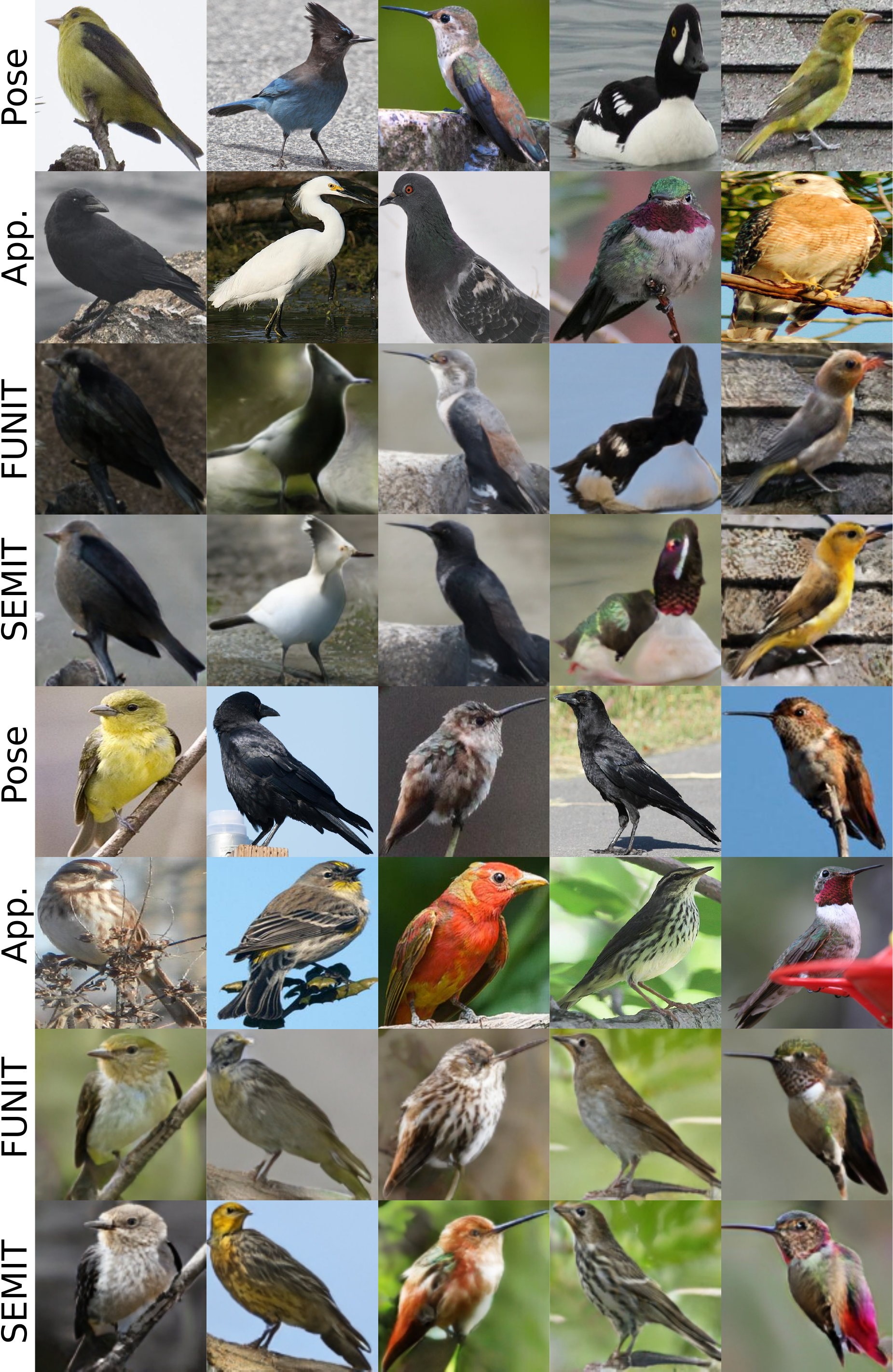}\vspace{-1mm}
   \caption{\small Qualitative comparison on the Birds
   datasets.}
   \label{fig:single_dataset_birds}\vspace{-4mm}
\end{figure*}

\begin{figure*}
    \centering
   \includegraphics[angle=90,width=0.95\textwidth]{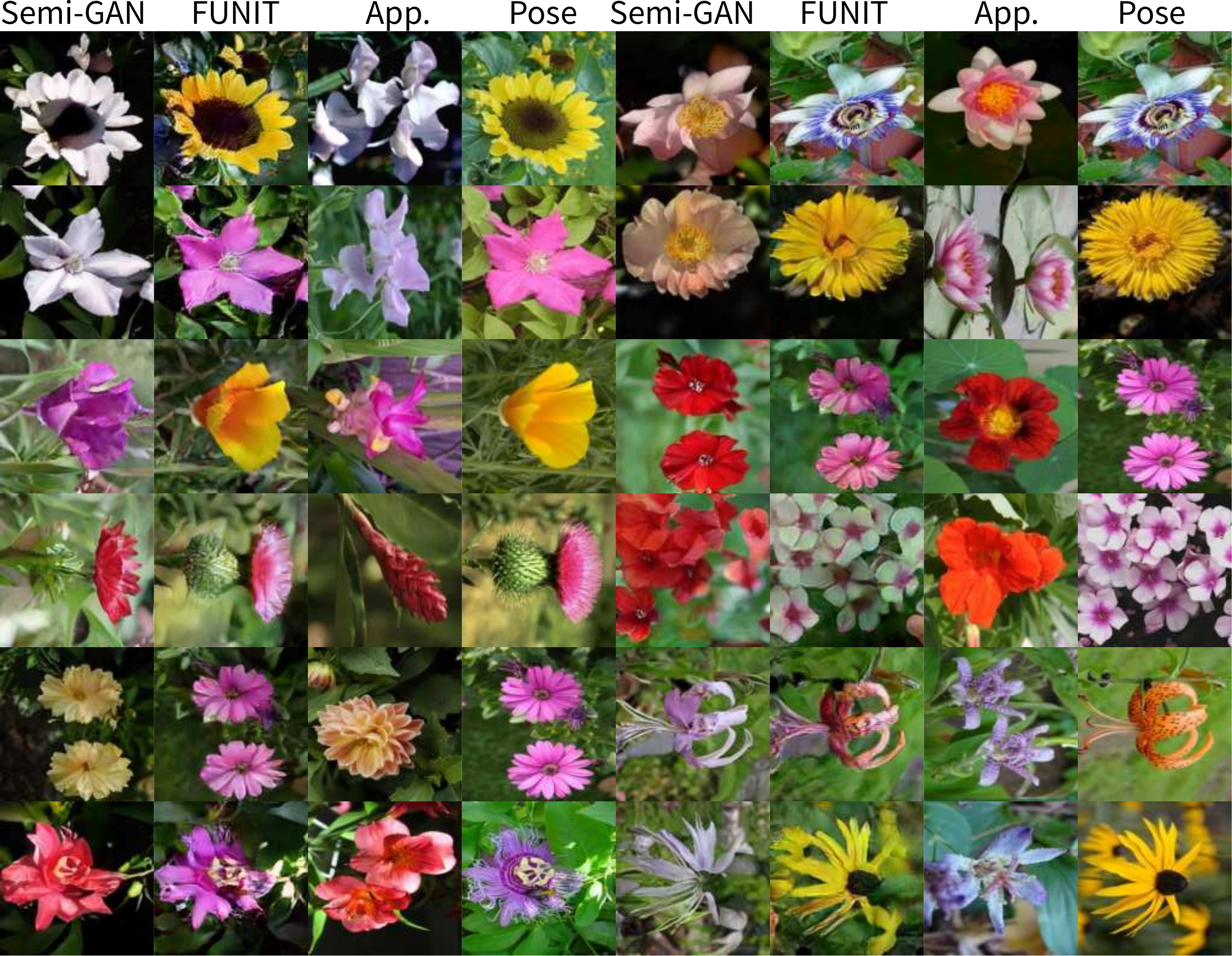}\vspace{-1mm}
   \caption{\small Qualitative comparison on the Flowers
   datasets.}
   \label{fig:single_dataset_flowers}\vspace{-4mm}
\end{figure*}

\begin{figure*}
    \centering
   \includegraphics[angle=90,width=0.90\textwidth]{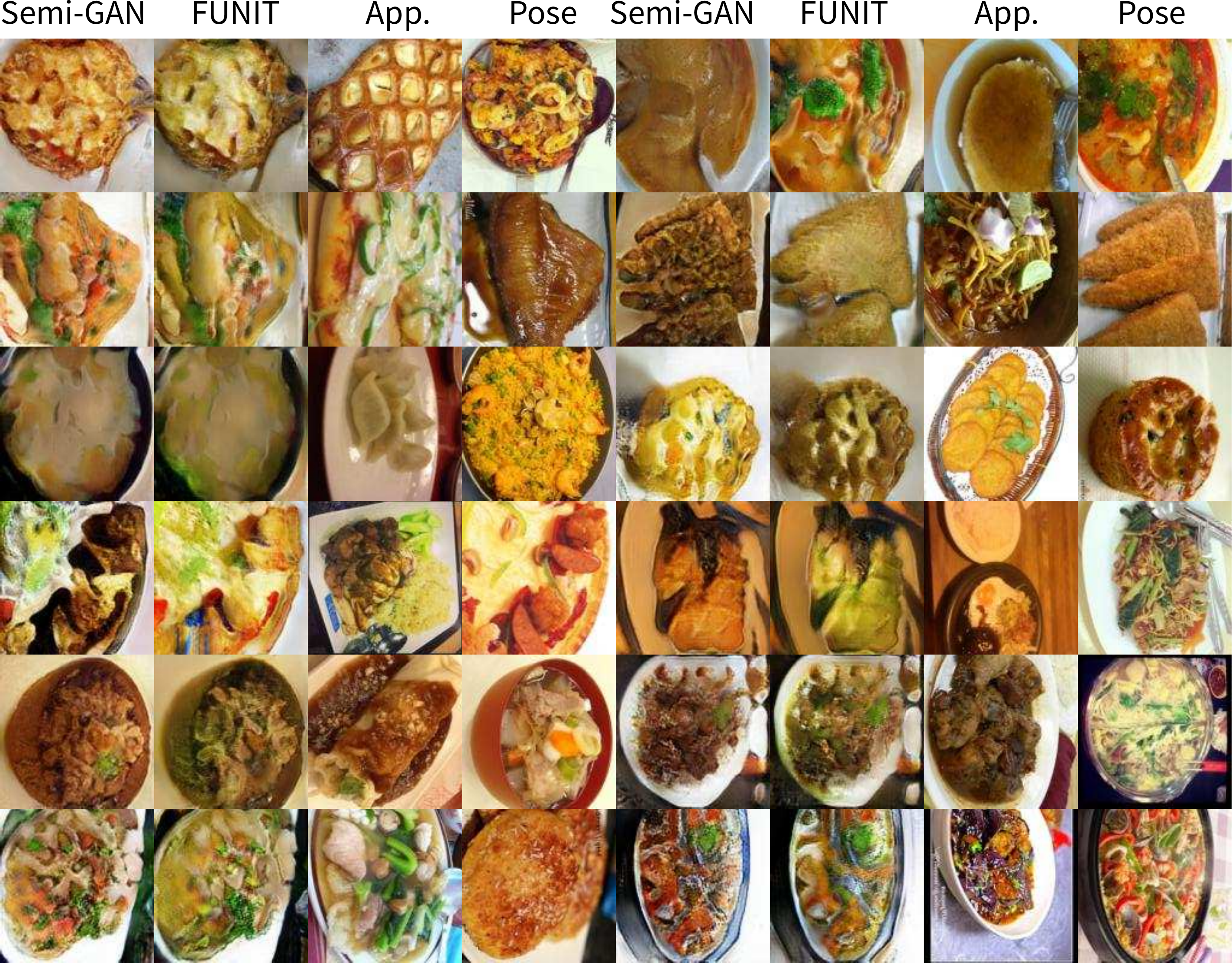}\vspace{-1mm}
   \caption{\small Qualitative comparison on the Foods
   datasets.}
   \label{fig:single_dataset_foods}\vspace{-4mm}
\end{figure*}

We provide additional results for models trained on a single dataset in  Figs.~\ref{fig:single_dataset_animals},~\ref{fig:single_dataset_birds},~\ref{fig:single_dataset_flowers},~\ref{fig:single_dataset_foods}.

\section{Qualitative results for models trained on multiple datasets} 
\begin{figure*}
    \centering
   \includegraphics[width=0.95\textwidth]{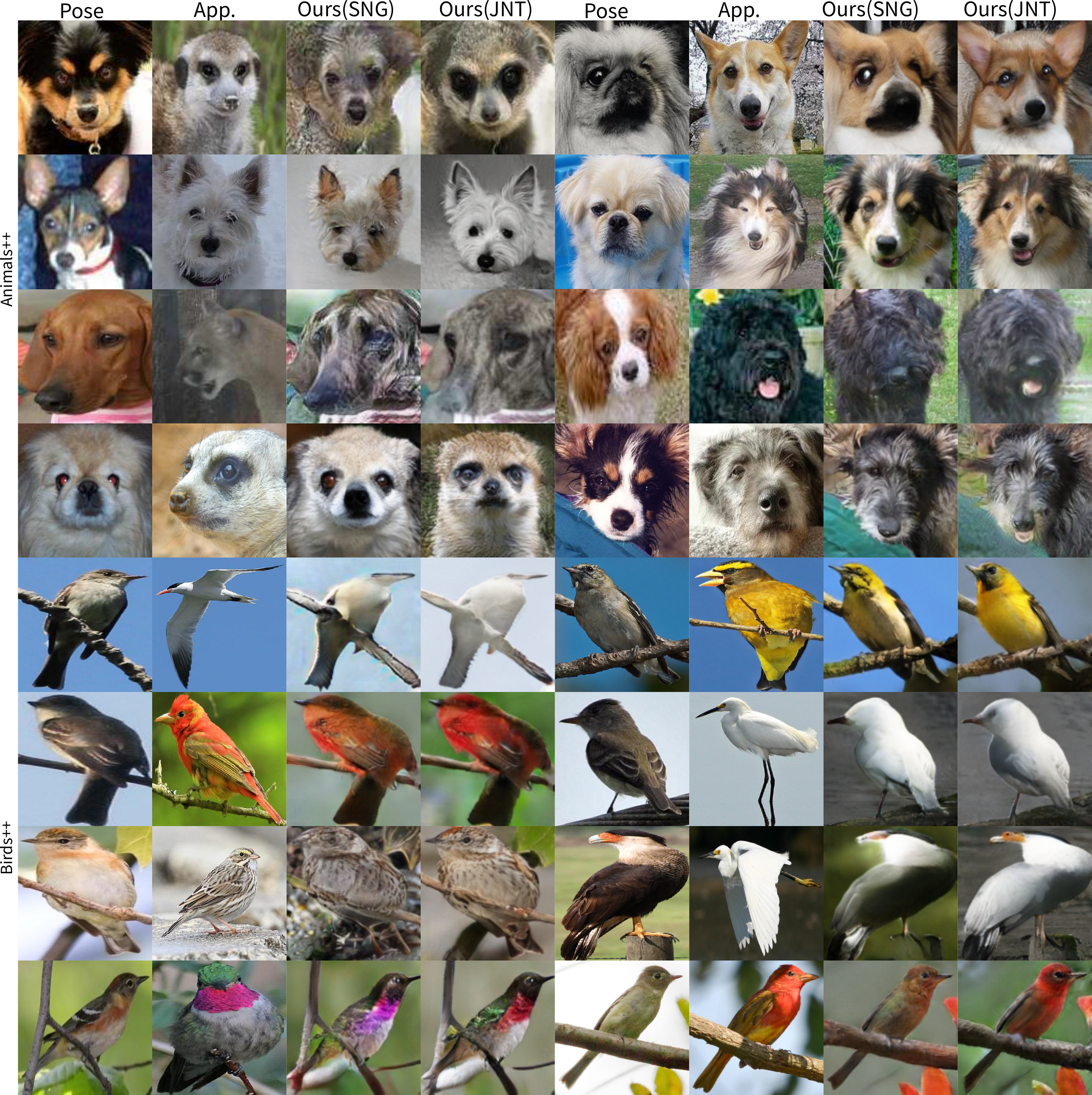}\vspace{-1mm}
   \caption{\small Results of our method on a single dataset (SNG) and joint datasets (JNT).}
   \label{fig:multi_dataset_flowers}\vspace{-4mm}
\end{figure*}

We provide additional results for models trained on multiple datasets in Fig.~\ref{fig:multi_dataset_flowers}.

\section{Quantitative results for models trained on multiple datasets} 
We provide additional quantitative results for models trained on multiple datasets in Tabs.~\ref{tab: results_animals} and ~\ref{tab: results_birds}.

\end{document}